\documentclass{article}

 \usepackage[eandd, final]{neurips_2026}

\usepackage[utf8]{inputenc} 
\usepackage[T1]{fontenc}    
\usepackage{hyperref}       
\usepackage{url}            
\usepackage{booktabs}       
\usepackage{amsfonts}       
\usepackage{nicefrac}       
\usepackage{microtype}      
\usepackage{xcolor}         

\title{Efficient Benchmarking Is Just Feature Selection and Multiple Regression}

\author{%
  Sam Bowyer\thanks{Work done during an internship at Cohere.}\\
  Cohere, University of Bristol\\
  \texttt{sam.bowyer@cohere.com} \\
  \And 
  Acyr Locatelli \\
  Cohere \\
  \texttt{acyr@cohere.com}
  \And
  Kris Cao \\
  Cohere \\
  \texttt{kriscao@cohere.com} \\
}

\usepackage{amsmath}
\usepackage{bm}
\usepackage{graphicx} 
\usepackage{subcaption}
\usepackage[ruled,vlined]{algorithm2e}
\usepackage{longtable}
\usepackage{booktabs}
\usepackage{tabularx} 
\newcolumntype{L}{>{\raggedright\arraybackslash}X}

\usepackage[sort&compress]{cleveref}
\crefname{section}{§}{§§}
\Crefname{section}{§}{§§}
\crefname{figure}{Fig.}{Figs.}
\Crefname{figure}{Fig.}{Figs.}
\crefname{equation}{Eq.}{Eqs.}
\Crefname{equation}{Eq.}{Eqs.}
\crefformat{equation}{Eq.~(#2#1#3)}
\crefrangeformat{equation}{Eqs.~(#3#1#4) to~(#5#2#6)}
\crefmultiformat{equation}{Eqs.~(#2#1#3)}{ and~(#2#1#3)}{, (#2#1#3)}{ and~(#2#1#3)}

\newcommand{\M}[0]{\mathcal{M}}
\newcommand{\D}[0]{\mathcal{D}}
\newcommand{\C}[0]{\mathcal{C}}
\newcommand{\bfy}[0]{\mathbf{y}}
\newcommand{\bfq}[0]{\mathbf{q}}
\newcommand{\bfs}[0]{\mathbf{s}}
\newcommand{\bfw}[0]{\mathbf{w}}
\newcommand{\bfx}[0]{\mathbf{x}}
\newcommand{\bfX}[0]{\mathbf{X}}
\newcommand{\bfI}[0]{\mathbf{I}}
\newcommand{\bfalpha}[0]{\bm{\alpha}}

\newcommand{\mtest}[0]{m^*}

\DeclareMathOperator*{\argmax}{arg\,max}
\DeclareMathOperator*{\argmin}{arg\,min}

\begin{document}

\maketitle

\begin{abstract}
    Efficient benchmarking techniques aim to lower the computational cost of evaluating LLMs by predicting full benchmark scores using only a subset of a benchmark's questions.
    By reframing this problem as an instance of \textit{multiple regression with feature selection}, we find that existing efficient benchmarking methods can be greatly improved by simply using kernel ridge regression at the prediction stage.
    Additionally, using an information-theoretic feature-selection algorithm called \textit{minimum redundancy maximum relevance} (mRMR), we can further improve upon these methods by selecting question subsets that will be maximally useful for prediction.
    Except in very data-poor settings, these approaches consistently achieve smaller prediction errors (in both MAE and RMSE), and greater ranking correlation between predicted and true scores (in both Spearman $\rho$ and Kendall $\tau$) across a range of benchmarks using both binary and continuous metrics.
    Furthermore, mRMR subsampling is much faster than competitor methods (which often involve fitting probabilistic models or running clustering algorithms), and is more likely to select the same questions under different random seeds or training data splits.
    Tutorial code can be found at \url{https://github.com/sambowyer/mrmr_eval}.
\end{abstract}


\section{Introduction}\label{sec:introduction}

Large language models (LLMs) continue to exhibit an incredible range of capabilities across a variety of tasks and domains. 
In order to accurately quantify, measure and interpret these capabilities, the need for rigorous evaluation is becoming increasingly important for both LLM developers and end-users. 
Moreover, as the size of these models increases, the corresponding inference costs also increase, particularly on complex, long-horizon benchmarks.
Traditional benchmarking involves evaluating an LLM on datasets containing roughly hundreds to thousands of examples (e.g. \citet{zellers_hellaswag_2019, hendrycks_measuring_2021, chen_evaluating_2021, liang_holistic_2023, fourrier_open_2024}), but the cost of evaluating larger, more complicated benchmarks, or even many benchmarks at the same time, can quickly become prohibitive.
Therefore, a recent line of work \citep{vivek_anchor_2024, maia_polo_tinybenchmarks_2024, kipnis_metabench_2025, balkir_confident_2026} investigates more efficient ways to measure model performance, primarily by sub-selecting particularly informative evaluation examples using data-driven approaches that make use of existing model evaluations.

As an example use-case, consider model training.
Ideally, a researcher would be able to evaluate every checkpoint of a training run on dozens of benchmarks in order to track the development of a variety of capabilities and skills that they would like a model to learn.
However, this not only takes up a lot of time---delaying any informed changes to the training procedure---but also takes up a considerable amount of compute that could otherwise be used on the training itself.
This problem becomes particularly acute if each benchmark question involves a complicated eval procedure, such as code execution \citep{jimenez_swe-bench_2024}, multi-turn conversation \citep{zhang_turnbench-ms_2025}, chain-of-thought reasoning steps \citep{wei_chain--thought_2022}, or multiple-answer generation (for example to compute a pass@$k$ metric \citep{chen_evaluating_2021}---for which low-variance estimates require many more than $k$ generations).
Even once training is complete, a researcher or user may want to run many ablations where they can vary inference-time options, such as sampling hyperparameters \citep{holtzman_curious_2020, renze_effect_2024, minh_turning_2025}, prompt formats \citep{wei_chain--thought_2022, sclar_quantifying_2024, romanou_brittlebench_2026} or the number of few-shot examples \citep{brown_language_2020}. 

\begin{figure}[!tbp]
    \centering
    \includegraphics[width=\textwidth]{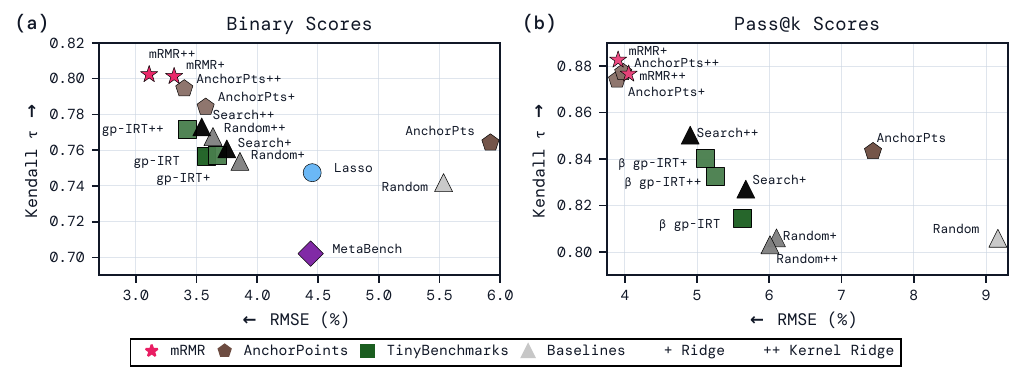}
    \caption{
    Our method (mRMR+/++) leads to significantly better benchmark score predictions in terms of reduced prediction error (RMSE) and increased ranking correlation (Kendall $\tau$).
    Using kernel ridge regression also improves the performance of existing efficient benchmarking methods AnchorPoints and TinyBenchmarks.
    \textbf{(a)} Benchmarks with binary scores: 5\% coresets and $M=30$ source models. \textbf{(b)} Benchmarks with pass@$k$ scores: 5\% coresets with $M=15$ source models.
    }
    \label{fig:pareto_main}
\end{figure}

\paragraph{Our approach}
Let $\mathcal{D}$ be a benchmark dataset containing $|\mathcal{D}|=N$ questions, and let $s(m,q_i)$ be the score achieved by model $m$ on question $q_i$, $i \in [N]$.
Given per-question results from $M$ \textit{source-models}, $\{s(m,q_i)\}_{m \in [M], i \in [N]}$, we define the task of efficient benchmarking in two stages: 
\begin{enumerate}
    \item[Stage 1:] Select a \textit{coreset} of questions $\C \subset \D$ of size $|\C| = n < |\D| = N$.
    \item[Stage 2:] Learn a function $f_\C: \mathbb{R}^n \to \mathbb{R}$ that predicts the full-benchmark score on test model $\mtest$ using only the coreset question scores: $f_\C([s(\mtest, q_i)]_{q_i \in \mathcal{C}})) \approx \bar{s}(\mtest, \D) = \frac{1}{N}\sum_{i=1}^N s(\mtest,q_i)$.
\end{enumerate}
The main conceptual step we propose is to reframe the problem of efficient benchmarking as an instance of a classic statistical problem: regression with feature selection.
Table \ref{tab:method_comparison} shows the approaches taken to both stages by various methods.
In particular, for stage one, we look to the feature-selection literature \citep{barbieri_analysis_2024, theng_feature_2024} and investigate the use of a popular feature selection technique known as \textit{Minimum Redundancy Maximum Relevance} (mRMR; \citet{peng_feature_2005, ding_minimum_2005, zhao_maximum_2019}).
For stage two, we primarily focus on ridge regression and polynomial kernel ridge regression \citep{hoerl_ridge_1970, saunders_ridge_1998, hastie_elements_2001}, the latter of which allows us to effectively predict benchmark scores using scores of both individual questions and combinations of questions.
These also have the benefit of naturally extending to both benchmarks with binary scoring (correct/incorrect) and continuous scoring (such as ROUGE-L \citep{lin_automatic_2004} or pass@$k$).

Thanks to the simplicity of mRMR and (kernel) ridge regression, our method is fast and easily adaptable to both discrete and continuous metrics.
As can be seen in \cref{fig:pareto_main} (a), using $M = 30$ source models on binary-valued benchmarks to construct and predict with coresets 5\% the size of the original dataset, our method not only predicts with low error (RMSE), but also maintains high correlation between predicted rankings of models and ground truth rankings (Kendall's $\tau$).
In \cref{fig:pareto_main} (b) we see that our method is still strong for continuous metrics, specifically, predicting pass@$k$ with $k \in \{1,2,4,\ldots,64\}$ on 5\% coresets constructed using pass@1 scores from $M=15$ models.
Interestingly, applying polynomial kernel ridge regression to the existing AnchorPoints method \citep{vivek_anchor_2024} also achieves comparable results, improving upon its original weighted mean predictions.
Furthermore, we find that coreset construction with mRMR is also \textit{stable} in the questions it selects, in the sense that it is more likely than competitor methods to select the same questions upon repeated runs with varying source model data (see \cref{fig:stability_and_difficulty_combined}). 

\section{Methods}\label{sec:methods}

\begin{table}[t]
\caption{Comparison of efficient benchmarking methods by feature selection and prediction strategies.}
\label{tab:method_comparison}
\centering
\begin{tabularx}{\textwidth}{l l l}
\toprule
\textbf{Method} & \textbf{Stage 1: Feature Selection} & \textbf{Stage 2: Prediction} \\
\midrule
Anchor Points & K-Means clustering on $\mathbf{q}_i$ & Weighted mean \\
\addlinespace[0.5ex]
gp-IRT        & K-Means on inferred IRT parameters & Weighted mean \\
\addlinespace[0.5ex]
MetaBench     & Maximising Fisher info. with inferred IRT parameters 
& GAM \\
Lasso         & $L_1$ penalty & Linear model \\
Random+       & Random sampling & Ridge regression \\
Search+       & Cross-validated random search & Ridge regression \\
mRMR++        & mRMR & Kernel ridge regression \\
\bottomrule
\end{tabularx}
\end{table}

\subsection{mRMR: Minimum Redundancy Maximum Relevance}\label{sec:mrmr}


Assume that there are $N$ features $X_1, \ldots, X_N$ with which we can predict a target variable $Y$.
For a feature-set (or \textit{coreset}\footnote{It is common to refer to collections of data points as \textit{coresets} and collections of data features as \textit{feature sets} \citep{moser_coreset_2026}. This distinction blurs in our setting, so we use the terms interchangeably to denote subsampled collections of questions.}) $\C$, containing $|\C| = n \leq N$ features, we define its \textit{relevance} to be the average mutual information (MI) between each of its features and the target $Y$, and the \textit{redundancy} to be the average MI between each pair of features:
\begin{equation}
    \label{eq:rel_red}\text{Relevance}(\C) = \frac{1}{n} \sum_{X_i \in \C} I(X_i, Y), \qquad
    \text{Redundancy}(\C) = \frac{1}{n(n-1)} \sum_{X_i \in \C} \sum_{X_j \neq X_i} I(X_i, X_j).
\end{equation}
Mutual information measures the dependence between two variables, and can be expressed in terms of the entropy of the variables $H(\cdot)$ (with joint entropy $H(\cdot, \cdot)$ and conditional entropy $H(\cdot | \cdot)$):
\begin{equation}\label{eq:mi_entropy}
    I(X,Y) = H(X) - H(X | Y ) = H(Y) - H(Y | X) = H(X) + H(Y) - H(X,Y).
\end{equation}
Equivalently, the MI can be expressed in terms of a KL-divergence between the joint distribution of the two variables $P_{X,Y}$, and the outer-product distribution of their marginals $P_X\otimes P_Y$.
For two continuous random variables $X$ and $Y$ with sample spaces $\Omega_X$ and $\Omega_Y$, marginal densities $p(x)$ and $p(y)$, and joint density $p(x,y)$ this can be written as:
\begin{equation}\label{eq:mi}
    I(X,Y) = D_{KL}(P_{X,Y}\; || \; P_X\otimes P_Y )= \int_{\Omega_X} \int_{\Omega_Y} p(x,y)\log\left(\frac{p(x,y)}{p(x)p(y)}\right)dydx \geq 0.
\end{equation}
The goal of mRMR is to construct a coreset $\C$ with maximal relevance and minimal redundancy.
Finding the globally optimal subset is NP-complete \citep{davies_np-completeness_1994}, so a greedy approach is usually taken.
Starting with a coreset containing the feature with maximal relevance $\mathcal{C}_{(1)} = \{\argmax_{X_i} I(X_i, Y)\}$, we iteratively add the feature that maximises an importance function $g$: 
\begin{equation}
    \C_{(t)} = \C_{(t-1)} \cup \argmax_{X_i \notin \C_{(t-1)}} g(X_i; \C_{(t-1)}, Y).
\end{equation}
This continues until we have a coreset $\C = \C_{(n)}$ of desired size $n$.
Many forms of $g$ exist (e.g. using F-statistics \citep{ding_minimum_2005} or random-forests \citep{zhao_maximum_2019};  see \cref{app:mrmr_objs} for more details), but the two most common are:
\begin{align}
    \text{(MI Difference)} \quad g^\text{MID}(X_i; \C_{(t-1)}, Y) =& I(X_i, Y) - \frac{1}{|\C_{(t-1)}|}\sum_{X_j  \in \mathcal{C}_{(t-1)}}I(X_j, X_i), \label{eq:mid} \\ 
    \text{(MI Quotient)} \quad g^\text{MIQ}(X_i; \C_{(t-1)}, Y) =& \left.I(X_i, Y) \middle/ \left(\frac{1}{|\C_{(t-1)}|}\sum_{X_j \in \mathcal{C}_{(t-1)}}I(X_j, X_i)\right).\right. \label{eq:miq}
\end{align}
In practice, we find that \textit{MIQ} leads to better downstream performance than MID or a redundancy-ignorant approach (see \cref{app:mrmr_objs}) for our use-case, and we therefore adopt it for our experiments.
We discuss the mRMR algorithm more in \cref{app:mrmr_algo} and \cref{app:coreset_rel_red}.

\subsection{Coreset construction via mRMR}\label{sec:mrmr_coresets}

To apply mRMR feature selection to efficient benchmarking on a dataset $\D$ of $|\D| = N$ questions, we simply need to decide how to represent the questions as features, and exactly what form the relevance target $Y$ should take.
Assuming we have per-question scores for $M$ \textit{source} LLMs, we represent each question as a vector containing per-model scores $\bfq_i = [s(m_1, q_i), \ldots, s(m_M, q_i)] \in \mathbb{R}^M$, $i \in [N]$,where $s(m,q)$ denotes the score achieved by model $m$ on question $q$.
For the relevance target $Y$, we simply use a vector containing the per-model summary-scores $\bar{s}(m, \D) = \frac{1}{|\D|}\sum_{q \in \D} s(m,q)$:
\begin{equation}
    \bar{\bfs} = [\bar{s}(m_1, \D), \ldots, \bar{s}(m_M, \D)] \in \mathbb{R}^M.
\end{equation}
Now we have $M$ datapoints for each question, $q_i$, and for the summary-score, $\bar{s}$, with which to estimate MI.
That is, we have $M$ samples from the joint space of $N+1$ random variables,
\begin{equation}
    s(\cdot \;, q_1),\; s(\cdot \;, q_2), \; \ldots, \; s(\cdot \;, q_N),\; \bar{s}(\cdot \;, \D).
\end{equation}
If the benchmark being considered only admits binary scores, $s(m,q) \in \{0,1\}$, then the redundancy part of our mRMR selection criteria (\cref{eq:mid} and \cref{eq:miq}) involves calculating the empirical MI between two binary random variables, $Q_i$ and $Q_j$,
\begin{align}\label{eq:mi-discrete}
    I(Q_i, Q_j) &= \sum_{q_i \in \{0,1\}} \sum_{q_j \in \{0,1\}} p(q_i,q_j)\log\left(\frac{p(q_i,q_j)}{p(q_i)p(q_j)}\right).
\end{align}
The necessary probabilities can be defined using the empirical proportions of zeros and ones across our $M$ models.
That is, for $a,b \in \{0,1\}$, and with the indicator function $\mathbb{I}$, we have:
\begin{align}
    &p(\mathbf{q}_i=a, \mathbf{q}_j=b) = \frac{1}{M} \sum_{m=1}^M \mathbb{I}(q_i^{(m)}=a)\mathbb{I}(q_j^{(m)}=b), \label{eq:binary_joint} \\
    &p(\mathbf{q}_i=a) = \frac{1}{M} \sum_{m=1}^M \mathbb{I}(q_i^{(m)}=a), \quad \text{ and } \quad p(\mathbf{q}_j=b) = \frac{1}{M} \sum_{m=1}^M \mathbb{I}(q_j^{(m)}= b). \label{eq:binary_marginals}
\end{align}
To calculate the relevance term involving $I(X_i, Y) = I(\bfq_i, \bar{\bfs})$, we require a discrete-continuous MI estimator, since the summary scores $\bar{\bfs}$ are not in general binary. 
We employ the estimator derived in \citet{ross_mutual_2014}, based on the continuous-continuous Kraskov–Stögbauer–Grassberger (KSG) estimator \citep{kraskov_estimating_2004}.
(We avoid simply discretizing our data and using a discrete-discrete MI estimator as this will give poor estimates unless $M$ is very large.)

In the case that a benchmark admits continuous scores, $s(m,q) \in \mathbb{R}$, we use a PCA-corrected KSG estimator \citep{gao_efficient_2015}.
Note that both this and the estimator above involve computing the $k$-nereast-neighbours for each datapoint. 
Therefore we have one hyperparameter $k \in \mathbb{N}$, which we search through $k=3,4,\ldots,9$, finding an optimal $k=5$ for binary data (\cref{sec:binary_experiments}) and $k=8$ for continuous data (\cref{sec:continuous_experiments}).  
(Further discussion on the MI estimators used in our experiments can be found in \cref{mrmr:app:mi_estimators}. Ablations on the value of $k$ and the standard non-PCA-corrected KSG estimator can be found in \cref{app:mrmr_objs}.) 

\subsection{Prediction using coreset results}\label{sec:prediction}


Once we have the coreset $\C = \{q_{\C_1}, \ldots, q_{\C_n}\}$, we must learn a function $f: \mathbb{R}^n \to \mathbb{R}$ which predicts the full benchmark score $\bar{s}(m, \D)$ using only the coreset-question scores $\{s(m, q_{\C_1}), \ldots, s(m, q_{\C_n})\}$.
Any suitable regression technique here should work, however, we proceed with ridge regression \citep{hoerl_ridge_1970, frank_statistical_1993, hastie_ridge_2020, hastie_elements_2001} since it is simple, quick and effective. 
(In \cref{app:relevance_target} we investigate the use of alternative regression methods.) 

\paragraph{Ridge Regression}
Given $m$ datapoints each with $n$ features $\mathbf{X} = [\bfq_1 \ldots \bfq_n] \in 
\mathbb{R}^{m \times n}$ and a target $\bfy = \bar{\bfs}$, ridge regression learns $w_0 \in \mathbb{R}, \bfw \in \mathbb{R}^n$ to minimise an $L_2$-penalised least squares objective:
\begin{align}\label{eq:ridge}
    \hat{\mathbf{w}} = \argmin_{\bfw, w_0}\nolimits \left\{||\bfy - (w_0 + \bfX \bfw)||_2^2 + \lambda ||\bfw||_2^2 \right\}
    = (\bfX^T\bfX + \lambda \bfI_n)^{-1}\bfX^T\bfy,
\end{align}
where $\lambda > 0$ is a regularisation coefficient (which we select via cross-validation) and $\bfI_m$ is the $m \times m$ identity matrix.
This gives us the prediction function $f(\bfx) = w_0 + \bfx^T \bfw$ for the per-question response vector of a test model $\bfx = [s(m, q_{\C_1}), \ldots, s(m, q_{\C_n})] \in  \mathbb{R}^n$.
The $L_2$ penalty shrinks coefficients and renders the matrix $(\bfX^T\bfX + \lambda\bfI)$ invertible, allowing for a closed-form solution.

\paragraph{Kernel Ridge Regression}
By allocating only a single weight per coreset question, standard ridge regression assumes that the summary score $\bar{s}$ is an additive linear function of individual question scores.
However, the performance gain signalled by correctly answering a pair of coreset questions may be greater (or less) than the sum of each question's individual contribution. 
To deal with this, we also investigate the use of \textit{kernel} ridge regression \citep{saunders_ridge_1998, hastie_elements_2001} with degree-$d$ polynomial kernels. 
Let $\bfx_i, \bfx_j \in \mathbb{R}^n$ be the $i$th and $j$th rows of $\bfX$, representing the score vectors for models $i$ and $j$ on the $n$ coreset questions.
We solve for dual weights $\bfalpha \in \mathbb{R}^m$ using the kernel matrix $K \in \mathbb{R}^{m \times m}$ where $[K]_{ij} = \kappa(\bfx_i,\bfx_j) = (\langle \bfx_i, \bfx_j\rangle + 1)^d$,
\begin{align}\label{eq:kernel_ridge}
    \hat{\bfalpha} = \argmin_{\bfalpha}\nolimits \{||\bfy - K\bfalpha||_2^2 + \lambda \bfalpha^TK\bfalpha\} 
    = (K+\lambda\bfI_m)^{-1}\bfy.
\end{align}
Thanks to the kernel trick (\citet{boser_training_1992}; see \cref{app:kernel_trick}) the degree-$d$ polynomial kernel allows us to predict $\bfy$ using the products of all groups of features up to size $d$ (e.g. $d=2$ lets us consider \textit{pairs} of features; and setting $d=1$ differs from regular ridge regression only in a rescaled bias term $w_0$).

\subsection{Other methods}\label{sec:other_methods}
\paragraph{Random} 
We sample $n$ questions from the benchmark independently at random without replacement, then take the mean score among these $n$ coreset questions as an estimate for the full benchmark score. 

\paragraph{Lasso} Replacing ridge regression's $L_2$ penalty with an $L_1$ penalty causes many of the coefficients $w_i$ to shrink to 0 \citep{tibshirani_regression_1996, hastie_elements_2001}. 
This gives an all-in-one coreset prediction strategy: we increase the regularisation strength $\lambda > 0$ until we only have $n$ nonzero coefficients.

\paragraph{Anchor Points}
\citet{vivek_anchor_2024} perform K-means clustering over $\bfq_i \in \mathbb{R}^M, i \in [N]$, and select the $K=n$ medoid questions to form $\mathcal{C}$.
To predict, coreset scores are combined in a weighted mean using each medoid's cluster size (normalised to sum to 1).
(For conciseness, we omit the "predictor" variant of this in which every non-coreset question score is predicted via an ensemble of univariate regressions on each coreset-question as it underperforms the weighted-mean approach.) 

\paragraph{Item Response Theory (IRT)}
Much recent work has focused on using IRT \citep{lord_statistical_1968, baker_item_2004, van_der_linden_handbook_2018} to provide informative representations of benchmark questions and model abilities \citep{lalor_building_2016, rodriguez_clustering_2022, jo_what_2025}.  
The two-parameter IRT model represents the probability of model $m$ correctly answering question $i$ as:
\begin{equation}\label{eq:irt}
    p_{i,m} = \mathbb{P}(s(m, q_i) = 1 \; | \; \theta_m, \alpha_i, \beta_i) =  1/\left(1+\exp(-\alpha_i^T\theta_m + \beta_i\right),
\end{equation}
where $\theta_m, \alpha_i \in \mathbb{R}^p$ and $\beta_i \in \mathbb{R}$ are latent variables representing, respectively, model abilities, question difficulty and a bias term, fitted using MCMC or variational inference. 
TinyBenchmarks \citep{maia_polo_tinybenchmarks_2024} defines two IRT-based methods.
In the first, Performance-IRT (\textit{p-IRT}), we estimate $\theta_{\mtest}$ for a new test model $\mtest$ via maximum likelihood on \cref{eq:irt} with fixed $\hat{\alpha}_i, \hat{\beta}_i$, and obtain per-question success probabilities $p_{i,\mtest}$ that we use as stand-ins for the true question scores on non-coreset questions. 
The second method, Generalised p-IRT (\textit{gp-IRT}), predicts a convex combination of the p-IRT score and an AnchorPoints score obtained via K-Means on the fitted mean latents $(\hat{\alpha}_i, \hat{\beta}_i) \in \mathbb{R}^{p+1}$ to get normalised cluster-size weights $\{w_i\}_{i: \;q_i \in \mathcal{C}}$:
\begin{equation}\label{eq:gpirt}
    f_\mathcal{C}^{\text{gp-IRT}}([s(\mtest, q_i)]_{q_i \in \mathcal{C}}) = \lambda \sum_{i: \; q_i \in \mathcal{C}} \nolimits w_i s(\mtest, q_i) + (1-\lambda) f_\mathcal{C}^{\text{p-IRT}}([s(\mtest, q_i)]_{q_i \in \mathcal{C}}).
\end{equation}
Here $\lambda \in [0,1]$ is chosen via a heuristic depending on the sample variance of all question scores across all source models, $\hat{\sigma}^2$, and an approximation of the IRT prediction bias.
The gp-IRT method tends to produce smaller errors than both the AnchorPoints method and p-IRT, so we omit p-IRT results.\footnote{Omitted results are available at \url{https://github.com/sambowyer/mrmr_paper}.}
However, the IRT model only supports binary metrics and running IRT inference is computationally expensive and leads to high variance-predictions (\cite{madaan_quantifying_2025}), hurting ranking prediction quality.
To adapt gp-IRT to continuous data, we use the $\beta$-IRT model from \citet{noel_beta_2007}, described in \cref{app:cirt}.
(See \cref{app:irt_dimension} for ablations on dimensionality $p$ and other continuous IRT models.)

\paragraph{Metabench} \citet{kipnis_metabench_2025} propose an IRT-based method for constructing a coreset from amongst various different benchmarks. 
After fitting an IRT model, as in \cref{eq:irt}, the authors split the range of fitted abilities $\theta_i$ into $n$ quantile bins and for each bin select the most informative question (in terms of Fisher information). 
Prediction happens on coreset questions using a linear GAM.
This method requires a large $M$ to get a good model fit and so tends to underperform in our setting.

\paragraph{Refitting Ridge}
We also experiment with alternative versions of AnchorPoints and gp-IRT methods in which we learn a (kernel) ridge predictor on their pre-constructed coresets.
We denote these methods with an appended `+' for standard ridge regression and a `++' appended when using kernel ridge regression with $d=2$.
(We investigate higher values of $d$ in \cref{app:kernel_degree} and find that $d=2$ performs the best.)
For example, `AnchorPoints+' and `AnchorPoints++' refer to using ridge and kernel ridge regression respectively to make predictions with the coreset chosen by AnchorPoints.

\paragraph{Exhaustive Random Search}
Finally, we look at a hard baseline labelled \textit{Search+} (and \textit{Search++}), where we sample 10,000 coresets at random, use 75\% of the evaluation models to fit a ridge regressor, and test the regressor on the remaining 25\%.
We then select the coreset with the lowest test MAE and refit the regressor using the full model pool before computing the test error on the test models. 

Note that both Search+ and Random+ were proposed as strong baselines in \citet{zhang_how_2025}.
We find that they can be beaten using more principled coreset construction and kernel ridge regression.

\begin{figure}[!tbp]
    \centering
    \includegraphics[width=\textwidth]{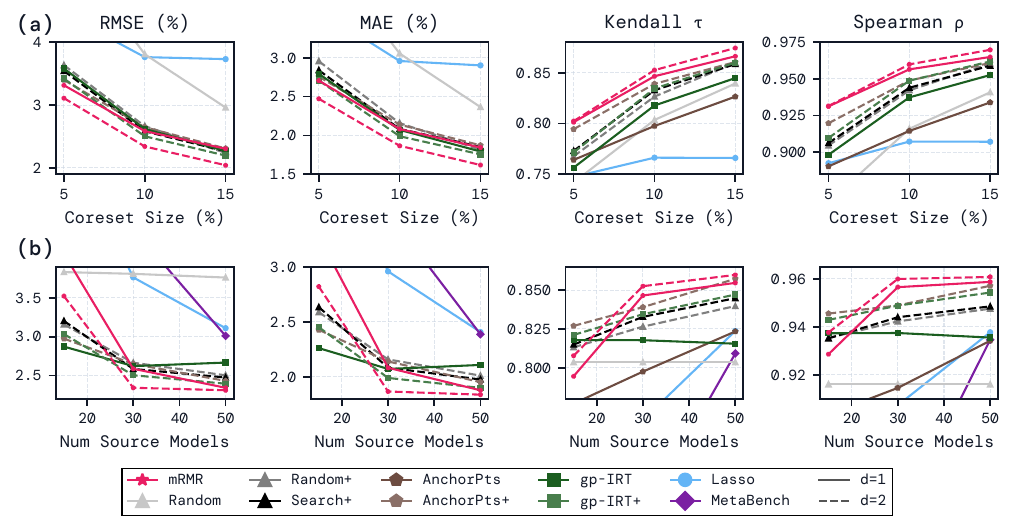}
    \caption{
    (\textbf{a}) With $M=30$, mRMR++ consistently achieves lower predictive errors and higher ranking correlations than other methods at a variety of coreset sizes.
    (\textbf{b}) Whilst mRMR methods struggle with very small $M$, for $M \in \{30,50\}$, mRMR++ dominates in both errors and ranking correlations using 10\% coresets. (For clarity, we omit $d=1$ methods here except for mRMR to show the relative size of underperformance.)
    }
    \label{fig:binary_results}
\end{figure}

\begin{figure}[!tbp]
    \centering
    \includegraphics[width=\textwidth]{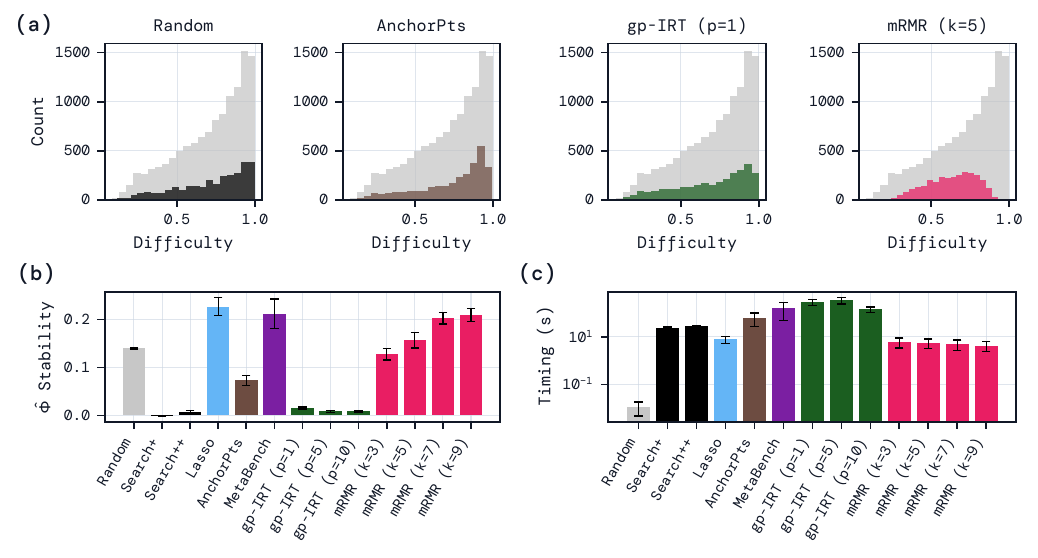}
    \caption{
    (\textbf{a}) Question difficulty distribution (proportion of all models which fail a given question) over coresets (coloured) vs full benchmarks (grey). 
    (MMLU, other datasets are shown in \cref{app:coreset_difficulty}.)
    (\textbf{b}) mRMR achieves greater coreset stability across random seeds, $M$ and coreset size compared to all other competitive methods.
    (\textbf{c}) mRMR is significantly faster than all others except random sampling. 
    }
    \label{fig:stability_and_difficulty_combined}
\end{figure}



\section{Experiments}\label{sec:experiments}

\paragraph{Datasets}

For binary metrics (\cref{sec:binary_experiments}), we take results from two online open-source leaderboards as processed by \citet{zhang_how_2025}: Open-LLM-Leaderboard \citep{fourrier_open_2024} which contains 448 models evaluated on 7 benchmarks, and HELM-Lite \citep{liang_holistic_2023} which contains between 79 and 83 models evaluated on six benchmarks.
For non-pass@$k$ continuous metrics (\cref{sec:continuous_experiments}) we take eval data from \citet{balkir_confident_2026} on two document summarisation datasets: GovReport \citep{huang_efficient_2021} and BioLaySumm \citep{xiao_overview_2025}.
For GovReport we have two metrics, ROUGE-L \citep{lin_automatic_2004} and BERTScore \citep{zhang_bertscore_2020}. For BioLaySumm we additionally have FKGL scores \citep{kincaid_derivation_1975}, giving us a total of 5 dataset-metric combinations (with all metrics scaled to $[0,1]$) for 21 models each run at 4 temperatures. 
Finally, for pass@$k$ metrics (\cref{sec:pass@k_experiments}) we ran four different Python coding benchmarks: MBPP, \citep{austin_program_2021}, MBPP+ \citep{liu_is_2023}, LBPP \citep{matton_leakage_2024} and HumanEval \citep{chen_evaluating_2021} with 10 models, each evaluated at 3 temperatures with 128 generations per question. 
After filtering out model-temperature combinations in which more than 1\% of samples failed to produce 128 generations, this left us with 24 combinations for MBPP+ and HumanEval, and 25 for MBPP and LBPP.
We separate out these experiments from \cref{sec:continuous_experiments} in order to assess how well static coresets fare at predicting related metrics.
In particular, we use coresets generated on pass@1 results to predict results for pass@$k$ for $k \in \{1,2,4,8,16,32,64\}$. 

We ran each method with $M \in \{15, 30, 50\}$ in \cref{sec:binary_experiments}, $M \in \{16, 32, 52\}$ in \cref{sec:continuous_experiments} and $M \in \{5,10,15\}$ in \cref{sec:pass@k_experiments} leaving the rest of the models as a test set.
We repeat experiments using five random seeds to generate five train-test splits. To avoid contamination, we do not allow the train and test set to contain the same model with different temperatures.
The training (source) models are selected via stratified sampling over ten equally-spaced bins on the range of full benchmark scores in order to ensure that the variety of model behaviour is well represented.
In each case, we run experiments targeting coreset sizes 5\%, 10\% and 15\% of the original benchmark size.

\paragraph{Metrics} 
In each experiment setting we focus on four key metrics to assess the quality of prediction (for full details see \cref{app:metric_descriptions}).
Firstly we look to minimise prediction error, specifically root-mean-squared error (RMSE) and mean-absolute error (MAE). 
Since we would like to compare these errors over multiple datasets, each of which may have their own level of summary-score variance, we report normalised versions of both RMSE and MAE, in which we weight each dataset's contribution by $\sigma_d^{-1}/(\sum_{d'=1}^D \sigma_{d'}^{-1})$ where $\sigma_d$ is the standard deviation of summary scores on dataset $d$.

Often, having high quality comparisons between models is more useful than knowing exact benchmark scores.
Therefore we also look at two \textit{rank correlation coefficients}.
These lie in $[-1,1]$ and measure the agreement between predicted and true rankings of a set of test models. 
We look at the Spearman $\rho$ \citep{spearman_proof_1904} (the correlation between the two ranking vectors) and the Kendall $\tau$ \citep{kendall_new_1938} ($\tau = \tau^*$ corresponds to having $(1-\tau^*)/2$ pairs of models ranked the wrong way around). 

We also want to measure the \textit{stability} of our coresets---the extent to which each method chooses the same questions given different random seeds or training data. 
We use a stability metric $\hat{\Phi}$ from the feature-selection literature \citep{nogueira_stability_2018}, where $0$ corresponds to every feature having an equal probability of being selected, and $1$ corresponds to selecting the exact same coreset every time.


\begin{figure}[tbp]
    \centering
    \includegraphics[width=0.925\textwidth]{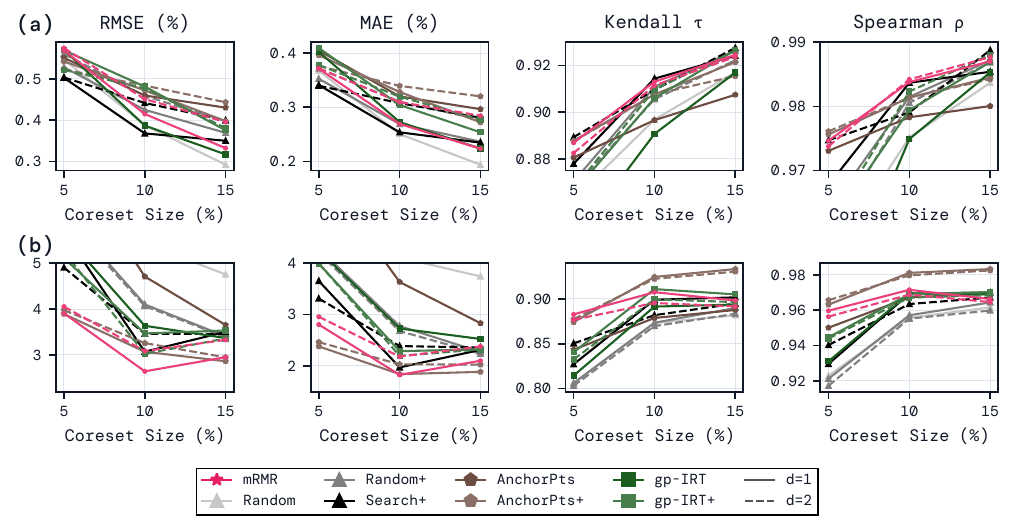}
    \caption{
    (\textbf{a}; $M=32$, see \cref{app:2_row} for other $M$ values) Continuous non-pass@$k$ benchmarks. 
    (\textbf{b}; $M=15$) Constructing coresets with pass@1 and predicting on all $k \in \{1,2,4,\ldots,64\}$ . 
    }
    \label{fig:continuous_results}
\end{figure}

    



\paragraph{Baselines} 
We compare our method against all the alternatives described in \cref{sec:other_methods}. 
In \cref{sec:continuous_experiments}, we also have to decide how methods should predict pass@$k$ with $k \neq 1$ using coresets generated from pass@1 data.
For methods which fit a regression, we can simply refit using the coreset questions.
Since AnchorPoints and gp-IRT use K-means to pick representative questions for their coreset, we use the same weighted sum for every value of $k$ (and we set $\lambda=1$ for gp-IRT since IRT predictions here are undefined).
Full implementation details of all methods can be found in \cref{app:experiment_details}. 

\subsection{Binary scores}\label{sec:binary_experiments}
For binary evaluations, \cref{fig:binary_results} shows that our method dominates other methods both in terms of minimising prediction error and maximising ranking correlation coefficients, particularly with quadratic kernels ($d=2$).
We see in \cref{fig:stability_and_difficulty_combined} (a) that gp-IRT and AnchorPoints tend to pick questions which are \textit{representative} of the entire dataset (due to their K-means approach), whereas mRMR focuses on medium-difficulty questions. 
Whilst this generally allows mRMR to select more \textit{informative} questions with greater stability (see \cref{fig:stability_and_difficulty_combined} (b)) and better predictive power, if $M$ is too small to identify the informative questions, then gp-IRT's broad, representative approach is the best option for low errors (although its low-stability and high-variance leads it to underperform in ranking predictions).
\cref{fig:stability_and_difficulty_combined} (c) shows that mRMR is orders of magnitude faster on binary data than all comparable methods.

\subsection{Continuous scores}\label{sec:continuous_experiments}

\cref{fig:continuous_results}(a) shows that for our non-pass@$k$ continuous metrics, mRMR struggles to find consistently better coresets than the brute-force approach of Search+, however, it does tend to outperform the more principled AnchorPoints and gp-IRT methods, particularly in terms of ranking correlations. 


\subsection{\texorpdfstring{Pass@$k$}{Pass@k} Generalisation}\label{sec:pass@k_experiments}
In \cref{fig:continuous_results} (b) we see that the coresets constructed by AnchorPoints, and to a lesser extent mRMR, on pass@1 scores lead to better generalisation in terms of predicting the other pass@$k$ scores.
Figure \ref{fig:stability_and_difficulty_combined} (a) shows mRMR selects questions of `average' difficulty. However, question difficulty shifts as the $k$ in pass@k changes, so what is `average' for one $k$ might not be for other $k$.
That being said, both these methods significantly outperform the others for 5\% coresets, and AnchorPoints+/++ is the clear winner for 10\% and 15\%.

Interestingly, we see in both continuous settings of \cref{fig:continuous_results} that regular kernel regression outperforms quadratic kernel ridge regression. 
Presumably this is because the continuous per-question scores already carry fine-grained information which can only be captured in the binary case by combining scores across multiple questions using e.g. polynomial kernels.
Thankfully both of these are extremely cheap and simple to implement, so it would be possible in practice to perform cross-validation to determine which method to use in a given situation.

\section{Discussion}\label{sec:limitations}
\begin{figure}[!tbp]
    \centering
    \includegraphics[width=\textwidth]{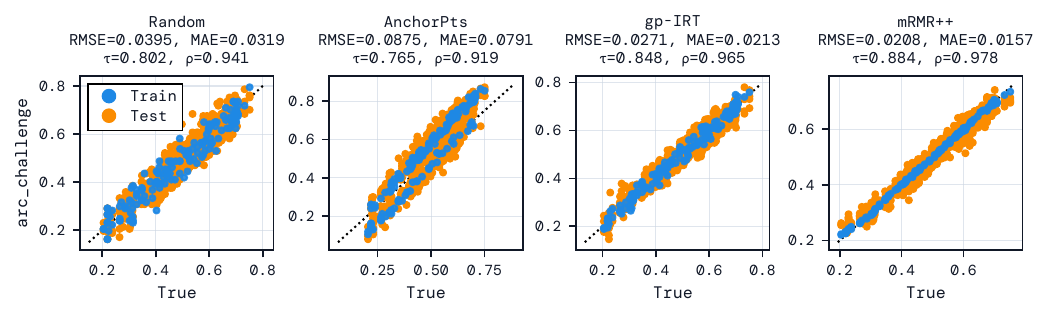}
    \caption{True vs. predicted summary scores on ARC-C for source (train) models and test models for all five random seeds and training data splits. 
    (Per-seed plot and more datasets are shown in \cref{app:true_v_pred}.)
    }
    \label{fig:true_vs_pred_arc-c}
\end{figure}

We've shown that ridge regression and kernel ridge regression work well in a range of settings and for a range of coresets.
\cref{fig:true_vs_pred_arc-c} shows true versus predicted summary scores for the methods Random, AnchorPoints, gp-IRT and kernel ridge (with mRMR coresets) on the ARC-C dataset.
This shows that kernel ridge achieves much lower training RMSE than the other methods (0.006 compared to 0.040 (Random), 0.081 (AnchorPoints), and 0.020 (gp-IRT)).
Interestingly we can also see the negative effect of AnchorPoints' low-stability and weighted-mean approach: two clear modes of training-point prediction behaviour representing very different coresets chosen in different data splits (see the per-seed version in \cref{fig:true_vs_pred_arc-c_per_seed}).
Whilst mRMR++ produces lower-variance predictions around the centre of performance compared to other methods, it has a tendency to over-predict at the floor of true model performance and under-predict at the ceiling, leading to a slight `S' shape, a common artifact of mean regression (this is more visible in other datasets, see \cref{fig:true_vs_pred_full}).
This suggests that our method's coresets should not be used to test models that are significantly worse or significantly better than the source models, which may fail or pass all questions in the coreset.
Finally, in \cref{app:ridge_underspecification} we discuss how the degree of underspecification (i.e. the ratio $M/n$; the number of datapoints per parameter) affects the performance of all discussed strategies.

\section{Related work}\label{sec:related_work}
We've shown that in very common and important efficient benchmark setting requiring \textit{static} coresets (for simplicity, interpretability and/or increased throughput via batching) and \textit{in-distribution} predictions (e.g. during model training or running non-frontier ablation experiments), we can leverage relatively simple feature selection and regression techniques to improve existing methods.
However, there is much recent work outside of this setting for which applying these techniques is not so straightforward.
\citet{zhang_how_2025} focuses on out-of-distribution efficient benchmarking---using a technique from  prediction-powered inference (PPI) methods \citep{angelopoulos_ppi_2023} to predict the scores of more powerful models using only data from less powerful source models.
We explore this regime in \cref{app:ood_extrap}, finding that the PPI approach is indeed best here, although it underperforms in the in-distribution setting with which we are most concerned. 
In \cref{app:ood_new_family} also examine out-of-distribution performance defined in a different sense---predicting the scores of model families whose members have been held out from the training set.
This tightens the gap between the best performing methods, however, mRMR++ remains generally optimal.

Similarly, there is much work on adaptive methods: dynamically selecting questions from the benchmark to suit each test model \citep{li_active_2025, hofmann_fluid_2025, yuan_beyond_2025, balkir_confident_2026}).
This has many attractive features, e.g. selecting more informative examples per model, and the potential for better use of compute through early stopping.
Another setting for efficient benchmarking involves utilising a wider range of data than just per-question per-model scores, whether that be model output distributions (e.g. per-class logits) \citep{rubinstein_disco_2025}, 
hand-crafted features \citep{saranathan_sublime_2025}, LLM activations \citep{wang_effieval_2025}, model generations \citep{helm2026queryefficient}, or the effect of in-context learning \citep{zhao_bento_2025}.
Other work has focused on different granularities of prediction, either using only summary scores \citep{papailiopoulos_you_2026}, or learning a model to predict individual question scores \citep{pacchiardi_100_2024}, or accounting for the cost of human evaluation of coreset questions \citep{zouhar_how_2025}. 
Finally, we note that an alternative approach to efficient benchmarking suggested by ONEBench \citep{ghosh_onebench_2025} involves aggregating a selection of questions across multiple benchmarks (potentially differing between test models) and synthesising an overall ranking of test models using a Plackett-Luce model \citep{plackett1975analysis, luce2005individual}. 
We explore this setting further in \cref{app:onebench}, noting that in our static-coreset setting with no missingness among coreset questions, the Plackett-Luce approach does not outperform the methods discussed in the main work.

\section{Conclusion}\label{sec:conclusion}
We suggest that a reframing of efficient benchmarking as multiple regression with feature selection can help researchers explore a range of already well-studied techniques, such as mRMR and kernel ridge regression.
These two techniques in particular beat every other method on binary-scored datasets.
Whilst improving mRMR's coresets on continuous valued datasets remains future work (whether that be with improved MI estimators for our relatively few-data regime, or alternative feature importance functions $g$), we show that simply applying kernel ridge regression to existing clustering based approaches works extremely well and lead to coresets that generalise to multiple related metrics.
This provides significant compute savings, and we hope that this work also opens up fruitful avenues for future efficient benchmarking research.

\begin{ack}

Sam Bowyer worked on this project primarily during an internship at Cohere, with additional work undertaken whilst supported by UKRI/EPSRC via the University of Bristol's COMPASS CDT. 

\end{ack}

\bibliographystyle{unsrtnat}
\bibliography{refs.bib}

\newpage 

\appendix

\section{Greedy mRMR algorithm}\label{app:mrmr_algo}

\begin{algorithm}[H]
\DontPrintSemicolon
\SetKwInOut{Input}{Input}\SetKwInOut{Output}{Output}

\Input{Dataset of features $\mathcal{D} = \{X_1, \dots, X_N\}$, Target variable $Y$, Desired subset size $n$}
\Output{Feature set/Coreset $\mathcal{C}$}

\BlankLine
\tcp{Step 1: Selection of the first feature (Maximum Relevance)}
$X^* \leftarrow \argmax_{X_i \in \mathcal{D}} I(X_i, Y)$\;
$\mathcal{C}_{(1)} \leftarrow \{X^*\}$\;

\BlankLine
\tcp{Step 2: Greedy selection of remaining $n-1$ features}
\For{$t = 2$ \KwTo $n$}{
    \BlankLine
    \tcp{Maximize the mRMR objective, g (MID or MIQ)}
    $X^* \leftarrow \argmax_{X_i \in \mathcal{D} \setminus \mathcal{C}_{(t-1)}} g(X_i; \mathcal{C}_{(t-1)}, Y)$\;
    
    \BlankLine
    $\mathcal{C}_{(t)} \leftarrow \mathcal{C}_{(t-1)} \cup \{X^*\}$\;
}

\Return{$\mathcal{C}_{(n)}$}\;
\caption{Greedy mRMR Feature Selection}
\label{alg:mrmr_greedy}
\end{algorithm}

\section{Ridge regression details}

\subsection{The kernel trick}\label{app:kernel_trick}
We consider polynomial kernels of degree $d$,
\begin{equation}\label{eq:app_kernel}
    \kappa(\bfx_i, \bfx_j) = (\langle \bfx_i, \bfx_j\rangle + c)^d,
\end{equation}
where $\bfx_i, \bfx_j \in \mathbb{R}^n$ contain coreset question scores for models $i$ and $j$ and $c=1$ is fixed. 
Using these in kernel ridge regression allows us to regress on the products of groups of features up to size $d$.
In our case, this means considering the scores on not just each individual questions, but also the products scores on all pairs of questions, all triples of questions, and so on.
This phenomenon is commonly referred to as the \textit{kernel trick}. 
To see how it works, it's helpful to start with the $d=2$ and $c=0$ case. 

Let $x_{i1} \in \mathbb{R}$ be the $k$th element of $\bfx_i$, i.e. the score of model $i$ on the $k$th coreset question.
If we first suppose that there are only two coreset questions, $n=2$, then expanding out \cref{eq:app_kernel} gives us
\begin{align}\label{eq:kernel_expansion}
    (\langle \bfx_i, \bfx_j\rangle)^2 &= (x_{i,1} x_{j,1} + x_{i,2} x_{j,2})^2 \\
    &= x_{i,1}^2 x_{j,1}^2 + x_{i,2}^2 x_{j,2}^2 + 2x_{i,1} x_{j,1}x_{i,2} x_{j,2} \\
    &= \langle\underbrace{(x_{i,1}^2, x_{i,2}^2, \sqrt{2}x_{i,1} x_{i,2})}_{\Phi(\bfx_i)}, \underbrace{(x_{j,1}^2, x_{j,2}^2, \sqrt{2}x_{j,1} x_{j,2})}_{\Phi(\bfx_j)} \rangle,
\end{align}
where $\Phi(\bfx_i) \in \mathbb{R}^3$ is a mapping of $\bfx_i$ to a higher dimensional feature space (the \textit{Reproducing Kernel Hilbert Space}, or RKHS) which contains not just the squares individual scores of both coreset questions, but also the product of their scores (scaled by a factor of $\sqrt{2}$).
If we used a nonzero constant $c$, e.g. $c=1$, then we'd obtain a mapping to a larger feature space which includes a constant and linear terms as well as the quadratic terms above:
\begin{align}\label{eq:kernel_expansion_c1}
    (\langle \bfx_i, \bfx_j\rangle + 1)^2 &= (x_{i,1} x_{j,1} + x_{i,2} x_{j,2} + 1)^2 \\
    &= x_{i,1}^2 x_{j,1}^2 + x_{i,2}^2 x_{j,2}^2 + 2x_{i,1} x_{j,1}x_{i,2} x_{j,2} + 2x_{i,1} x_{j,1} + 2x_{i,2} x_{j,2} + 1 \\
    &= \langle(x_{i,1}^2, x_{i,2}^2, \sqrt{2}x_{i,1} x_{i,2}, \sqrt{2}x_{i,2}, 1), (x_{j,1}^2, x_{j,2}^2, \sqrt{2}x_{j,1} x_{j,2}, \sqrt{2}x_{j,2}, 1) \rangle,
\end{align}
Note that if we increased the coreset size $n > 2$, then our $d=2$ quadratic expansion would still lead to a feature space that computes all pairwise interactions between coreset question scores, and as long as $c \neq 0$, we'd also have the linear terms (i.e. individual coreset question scores) and a bias term as well.
Furthermore, setting $d > 2$ would give us a feature mapping that computes all possible monomials up to degree $d$, e.g. triple-question-products for $d=3$ or pairs of squared question scores for $d=4$.
This flexibility leads to greater predictive performance in \cref{sec:binary_experiments}, however, it can lead to under- or over-fitting if $d$ is set too high. 
The effect of this is explored in \cref{app:kernel_degree}.

\subsection{Underspecification}\label{app:ridge_underspecification}
Ridge regression was originally developed to deal with underspecified regression problems, where we have fewer datapoints than parameters. causing ordinary least-squares (OLS) regression (with no regularisation penalty) to be unable to identify suitable coefficients $\hat{\bfw}$, since the matrix $\bfX^T\bfX$ (from \cref{eq:ridge} with $\lambda=0$) is not full-rank and therefore non-invertible.

\begin{figure}[!htbp]
    \centering
    \includegraphics[width=\textwidth]{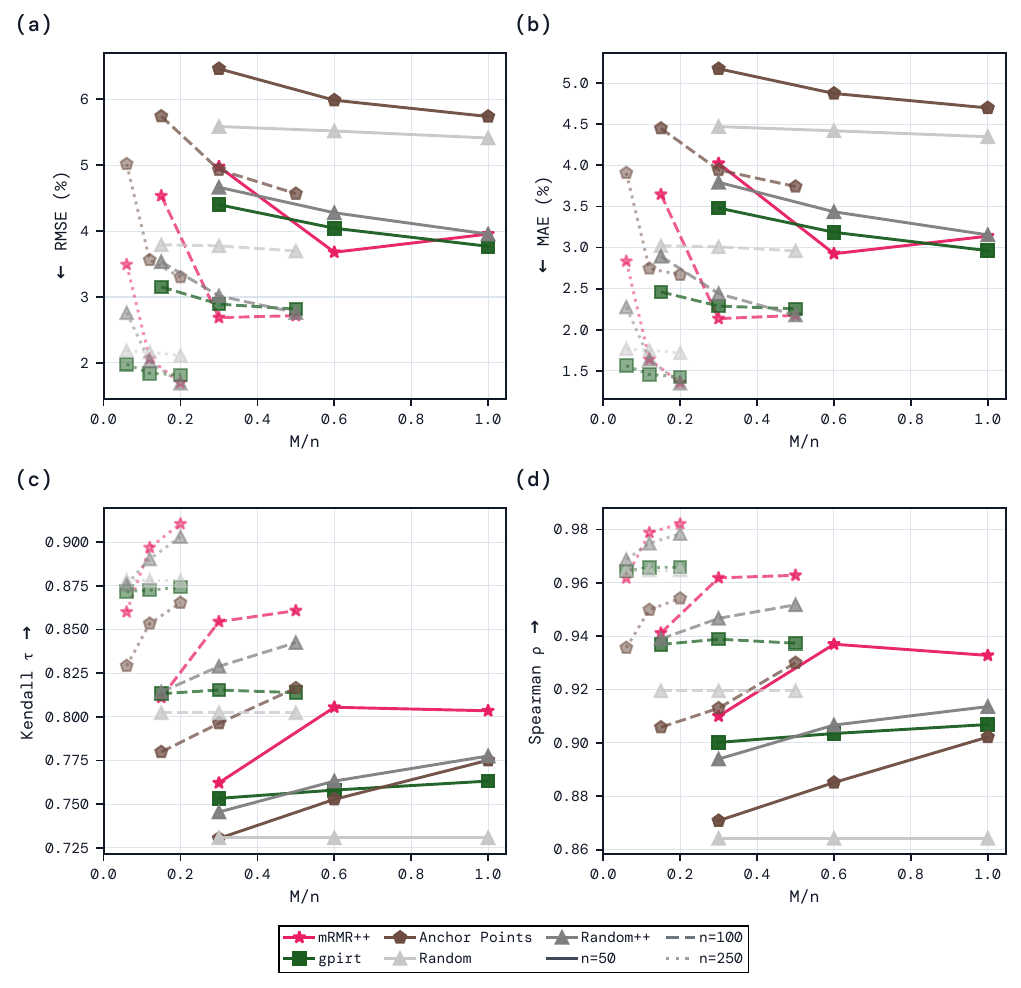}
    \caption{Performance gains from polynomial ridge regresion become less pronounced as the ratio $M/n$, the degree of underspecification, approaches zero. Here we use fixed coreset sizes $n \in \{50,100,250\}$ on binary datasets.}
    \label{fig:ridge_underspecification_fixedn}
\end{figure}

\begin{figure}[!tbp]
    \centering
    \includegraphics[width=\textwidth]{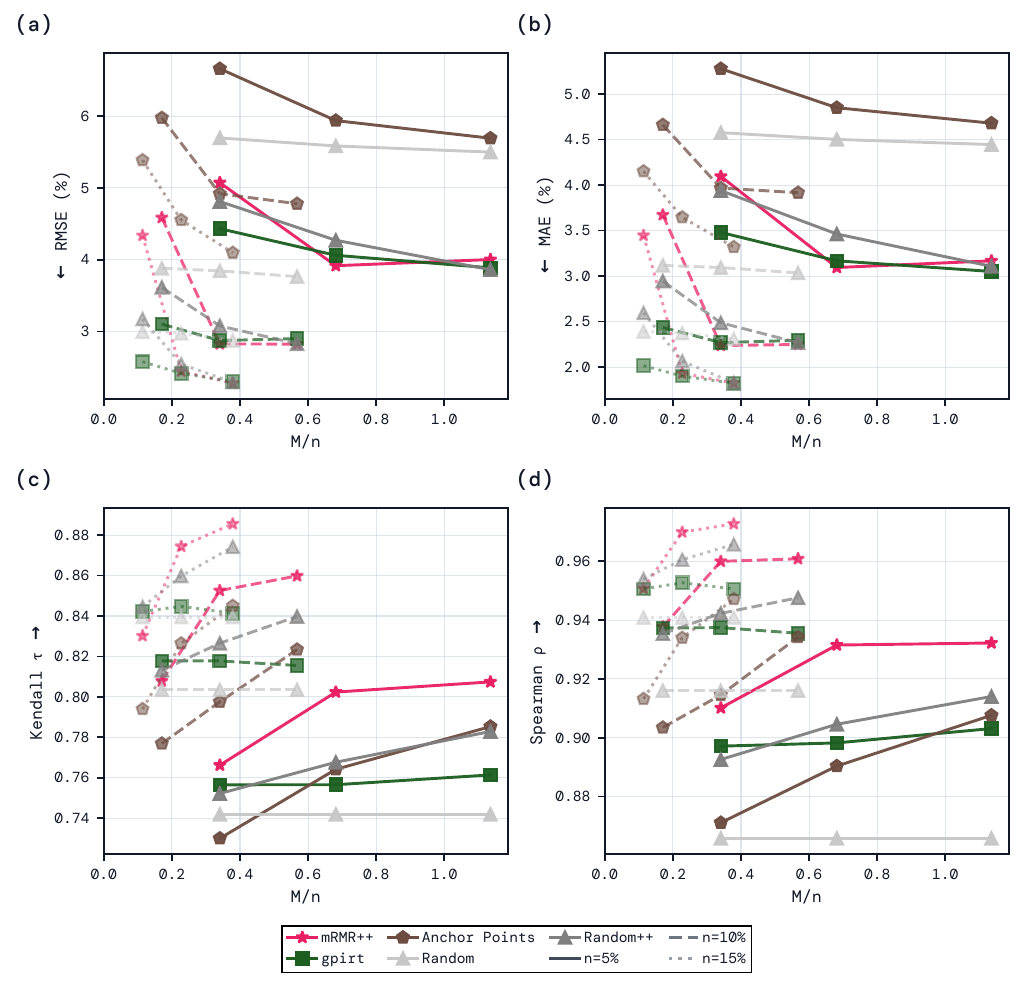}
    \caption{Same setting as \cref{fig:ridge_underspecification_fixedn} but with relative-sized coresets: $5\%, 10\%$ and $15\%$.}
    \label{fig:ridge_underspecification_percentn}
\end{figure}

In our setting, underspecification means we have fewer source models than coreset-questions, $M < n$. 
Whilst (polynomial) ridge regression \textit{can} deal with relatively small values of the ratio $M/n$, if this ratio grows too small then our signal-to-noise ratio among the coreset question scores can overwhelm and lead to poorer predictive performance.
This is shown in \cref{fig:ridge_underspecification_fixedn} where we've rerun the experiments in \cref{sec:binary_experiments} but instead target \textit{fixed coreset sizes} $n \in \{50, 100, 250\}$ rather than relative sizes of $5\%, 10\%$ and $15\%$.
Here we see that the relative performance gain from applying polynomial ridge regression (using both random and mRMR-constructed coresets) compared to gp-IRT and AnchorPoints decreases as the ratio $M/n$ gets closer to zero.
In \cref{fig:ridge_underspecification_percentn}, show the same results but on the $5\%, 10\%$ and $15\%$ relative-size coresets from \cref{sec:binary_experiments}.

\newpage
\subsection{Ridge degree comparison}\label{app:kernel_degree}
\begin{figure}[!tbp]
    \centering
    \includegraphics[width=\textwidth]{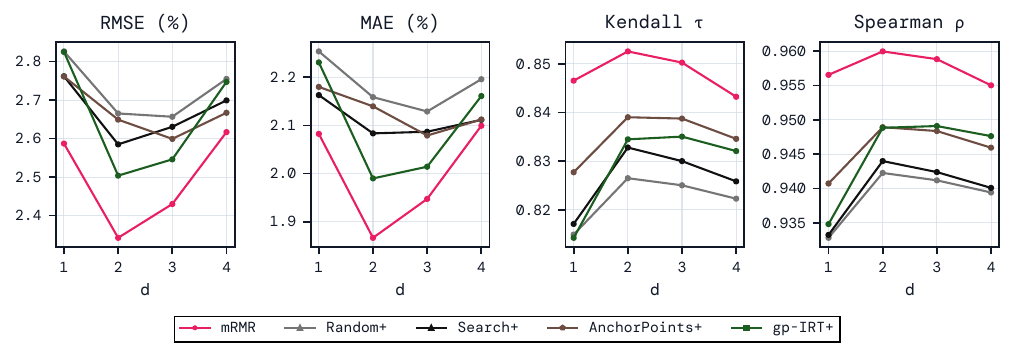}
    \caption{Binary-dataset results on 10\% coresets using $M=30$ source models. Quadratic kernels ($d=2$) tend to produce the best overall results for all methods, suggesting pairwise question combinations provide useful regression features without leading to under- or over-fitting.}
    \label{fig:ridge_degree_x_axis}
\end{figure}

In \cref{fig:ridge_degree_x_axis} we show the $M=30$, 10\% coreset-size results on binary datasets for methods using kernel ridge regression as we vary the polynomial kernel degree $d \in \{1,2,3,4\}$.
We observe that a quadratic kernel tends to produce optimal results for all considered methods, whilst higher degrees underperform, likely due to overfitting. 
Furthermore, we find that mRMR's coresets achieve the best RMSE, MAE, $\tau$ and $\rho$ compared to the other methods at any of these fixed $d$.


\newpage

\section{Metric descriptions}\label{app:metric_descriptions}
Here we provide more details on the metrics used to compare different efficient benchmarks in \cref{sec:experiments}.

\subsection{Errors}

Assume that we have a set of test models $\M'$ of size $M' = |\M'|$ with true benchmark scores $\bfy_\D = [\bar{s}(m'_1, \D), \ldots, \bar{s}(m'_{M'}, \D)] \in \mathbb{R}^{\M'}$ and predicted scores $\hat{\bfy}_{\D, f_\C} = [f_\C(m'_1), \ldots, f_\C(m'_{\M'})] \in \mathbb{R}^{\M'}$.
Firstly we consider two types of prediction error: mean absolute error (MAE; \cref{eq:mae}) and root mean squared error (RMSE; \cref{eq:rmse}):
\begin{align}
    \text{MAE}(f_\C, \D, \M') &= \frac{1}{|\M'|} \sum_{m=1}^{M'} \left|\hat{\bfy}_{\D, f_\C}^{(m)} - \bfy_\D^{(m)}\right|, \label{eq:mae} \\
    \text{RMSE}(f_\C, \D, \M') &= \sqrt{\frac{1}{|\M'|} \sum_{m' \in \M'} \left(\hat{\bfy}_{\D, f_\C}^{(m)} - \bfy_\D^{(m)}\right)^2}. \label{eq:rmse}
\end{align}
Since we want to look at the average error across multiple benchmarks, we aggregate errors over benchmarks $\D_1, \ldots, \D_D$ with a simple mean weighted by each benchmark's summary score standard deviation across test models (normalised across benchmarks):
\begin{align}
    \text{Agg-MAE}(f_\C, \D_{1:D}, \M') &= \sum_{d=1}^D \frac{\sigma_d^{-1}}{\sum_{d'=1}^D \sigma_{d'}^{-1}}\text{MAE}(f_\C, \D_d, \M'), \label{eq:agg-mae} \\
    \text{Agg-RMSE}(f_\C, \D_{1:D}, \M') &= \sum_{d=1}^D \frac{\sigma_d^{-1}}{\sum_{d'=1}^D \sigma_{d'}^{-1}}\text{RMSE}(f_\C, \D_d, \M'), \label{eq:agg-rmse}
\end{align}
where 
\begin{equation}
    \sigma_d = \sqrt{\frac{1}{M'}\sum_{m=1}^{M'}\left(\bfy_{\D_d}^{(m)} - \bar{\bfy}_{\D_d}\right)^2}
    \quad \text{ and } \quad
    \bar{\bfy}_{\D_d}= \frac{1}{|\M'|}\sum_{m=1}^{M'} \bfy_{\D_d}^{(m)}.
\end{equation}

\subsection{Rank correlation coefficients}

\paragraph{Pearson's \texorpdfstring{$r$}{r}}
\citep{pearson_note_1895} For two vectors $\bar{\bfs},  \hat{\bar{\bfs}} \in \mathbb{R}^M$ representing, respectively, ground-truth and predicted scored for $M$ models, the Pearson correlation is given by
\begin{equation}\label{eq:pearson}
    r(X,Y) = \frac{\text{Cov}(\bar{\bfs}, \hat{\bar{\bfs}})}{\text{StdDev}(\bar{\bfs})\text{StdDev}(\hat{\bar{\bfs}})} \in [-1,1],
\end{equation}
where $\text{Cov}(\bar{\bfs}, \hat{\bar{\bfs}})$ is the covariance between the variables $\bar{\bfs}$ and $\hat{\bar{\bfs}}$.
This tells us whether prediction \textit{values} are correlated with true \textit{values}, however, it will be affected by different models to different extents, depending on the magnitude of their summary scores (both predicted and true).

\paragraph{Spearman's \texorpdfstring{$\rho$}{rho}} \citep{spearman_proof_1904} Improves upon the Pearson correlation coefficient calculated between the rank variables $R_\D, \hat{R}_\D \in [M']^{M'}$, where the $i$-th element of $R_\D$ (resp. $\hat{R}_\D$) is the rank of the true (resp. predicted) score of the $i$-th model in $\M'$:
\begin{equation}\label{eq:spearman}
    \rho(\hat{R}_\D, R_\D) = \frac{\text{Cov}(\hat{R}_\D, R_\D)}{\text{StdDev}(\hat{R}_\D) \text{StdDev}(R_\D)} \in [-1,1].
\end{equation}
By looking at the ranks of models rather than their true values, we get a better sense of how effective an efficient benchmarking method is when comparing different models on the same benchmark, removing the dependence on the actual raw summary scores.

\paragraph{Kendall's \texorpdfstring{$\tau$}{tau}}
\citep{kendall_new_1938} Like Spearman's $\rho$, this also uses rank variables rather than raw summary scores. 
In particular, the Kendall $\tau$ of two rankings is computed as the number of pairwise comparisons which are concordant (in the same relative order in both rankings) minus the number of discordant pairs (in the opposite relative order), divided by the total number of such pairs, $\binom{n}{2}$:
\begin{equation}\label{eq:kendall_tau}
    \tau(\hat{R}_\D, R_\D) = \frac{1}{\binom{n}{2}} \sum_{1\leq m_1 < m_2 \leq M'} \text{sign}(\hat{R}_\D^{m_1} - \hat{R}_\D^{m_2})\text{sign}(R_\D^{m_1} - R_\D^{m_2}) \in [-1,1].
\end{equation}
Note that a value of $\tau$ means the proportion of incorrectly ranked pairs is $(1-\tau)/2$.

\subsection{Stability metrics}
We report two stability metrics designed to measure the extent to which coreset-construction methods select the same questions when presented with different numbers of source models, $M$, and different random seeds (which are also used to sample models for the train-test split).
Both metrics lie in $[0,1]$ with larger values indicating a greater tendency to select the same questions.
(In both, a stability of 1 is only possible if a method always selects the exact same coreset, however, a stability of 0 has different interpretations for each metric.)
We report aggregated stability metrics across different benchmarks by simply taking the mean. 

\paragraph{Noguiera's \texorpdfstring{$\hat{\Phi}$}{Phi}} \citep{nogueira_stability_2018}
Given a collection of $N$ features $\mathcal{D} = \{X_1, \ldots, X_N\}$, and $L$ feature-sets $\mathfrak{C} = \{\mathcal{C}^{(1)}, \ldots, \mathcal{C}^{(L)}\}$, we first compute the average size of the sets $\bar{n} = \frac{1}{L}\sum_{l=1}^L |\mathcal{C}^{(l)}|$ and for each feature $X_i$ we also calculate the proportion of the $L$ sets in which $X_i$ is present:
\begin{equation}
    \hat{p}_i = \frac{1}{L}\sum_{l=1}^L \mathbb{I}[X_i \in \mathcal{C}^{(l)}],
\end{equation}
as well as the unbiased sample variance of this feature's coreset-inclusion, $\hat{s}_i^2 = \frac{L}{L-1}\hat{p}_i(1-\hat{p}_i)$.

This frequency based approach aims to construct a metric which has expected value 0 under the null hypothesis $H_0$ that every choice of coreset $\mathcal{C} \in 2^\mathcal{D}$ is equally likely (which is the authors' definition of minimum stability).
In this case, it can be shown that the expected sample variance is given by
\begin{equation}
\mathbb{E}\left[\frac{1}{N}\sum_{i=1}^N \hat{s}_i^2 \; \Big| H_0\right] = \frac{\bar{n}}{N}\left(1-\frac{\bar{n}}{N}\right).
\end{equation}
Therefore the stability metric is defined as
\begin{equation}\label{mrmr:eq:noguiera}
    \hat{\Phi}(\mathfrak{C}) = 1 - \frac{\frac{1}{N}\sum_{i=1}^N \hat{s}_i^2}{\mathbb{E}\left[\frac{1}{N}\sum_{i=1}^N \hat{s}_i^2 \; \Big| H_0\right]} = 1 - \frac{\frac{1}{N}\sum_{i=1}^N \hat{s}_i^2}{\frac{\bar{n}}{N}\left(1-\frac{\bar{n}}{N}\right)}.
\end{equation}
This has many desirable properties, not only accounting for random coreset selection, but also being naturally robust to coresets of varying sizes. 




\paragraph{Hamming stability} \citep{dunne_solutions_2002} Rather than taking a frequency-based approach, it is also common to define stability using pairwise differences between feature-sets.
One very common such metric is the following, in which we represent a feature-set $\mathcal{C}$ as a binary vector $\mathcal{B}(\mathcal{C})$ of length $N$ such that the $i$th entry indicates whether $\mathcal{C}$ contains feature $X_i$:
\begin{equation}
\mathcal{B}(\mathcal{C})_i = 
\begin{cases} 
0, & X_i \notin \mathcal{C}, \\ 
1, & X_i \in \mathcal{C}.
\end{cases} 
\end{equation}
For a collection of $L$ feature-sets $\mathfrak{C}_n = \{\mathcal{C}^{(1)}, \ldots, \mathcal{C}^{(L)}\}$ all of size $n \leq N$, we define the \textit{Hamming stability for coresets of size $n$} to be one minus the average Hamming distance (equivalent to $L_1$ distance for binary vectors) between all pairs of feature-sets,
\begin{align}
H(\mathfrak{C}_n) &= 1- \frac{1}{\binom{L}{2}} \sum_{1 \leq l_1 < l_2 \leq L}  \frac{|| \mathcal{B}(\mathcal{C}^{(l_1)}) - \mathcal{B}(\mathcal{C}^{(l_2)}) ||_1}{2n} \\
&= 1 - \frac{1}{2n\binom{L}{2}} \sum_{1 \leq l_1 < l_2 \leq L}  \sum_{i=1}^N \mathbb{I}[X_i \in \mathcal{C}^{(l_1)}] \otimes \mathbb{I}[X_i \in \mathcal{C}^{(l_2)}],
\end{align}
where $\otimes$ represents the XOR operator.
In other words, this represents the average proportion of features which are shared between two feature-sets chosen from $\mathfrak{C}_n$.
To combine Hamming stability scores from collections of feature-sets of different sizes, $\mathfrak{C} = \{\mathfrak{C}_{n_1}, \ldots, \mathfrak{C}_{n_K}\}$, for integers $n_1, \ldots, n_K, K \in \mathbb{N}$ where
\begin{equation}
    |\mathcal{C}| = n_k \quad \forall \mathcal{C} \in \mathfrak{C}_{n_k},
\end{equation}
we simply take the mean Hamming scores across the coreset sizes, with each score inversely weighted by the size of its coreset collection $|\mathfrak{C}_{n_k}|$:
\begin{equation}\label{eq:hamming_different_sizes}
    H(\mathfrak{C}) = \frac{\sum_{k=1}^K H(\mathfrak{C}_{n_k}) / |\mathfrak{C}_{n_k}| }{\sum_{k=1}^K 1/|\mathfrak{C}_{n_k}|}.
\end{equation}
This helps maintain the interpretation that Hamming stability measures the average proportion of features which are shared between pairs from a collection of feature-sets.
(Since we run each experiment with five random seeds, we actually always have $|\mathfrak{C}_{n_1}| = \cdots = |\mathfrak{C}_{n_K}|$, so \cref{eq:hamming_different_sizes} reduces to a standard mean.)

\section{Experiment details}\label{app:experiment_details}

\subsection{Method implementation details}

For ridge regression and polynomial ridge regression methods, we select the regularisation coefficient $\lambda$, using leave-one-out cross validation over $\{0.1, 10^{-0.5}, 1, 10^{0.5}, 10\}$ for binary metrics and $\{0.01, 10^{-1.5}, 0.1, 10^{-0.5}, 1, 10^{0.5}, 10, 10^{1.5}, 100\}$ for the continuous (both pass@$k$ and non-pass@$k$), selecting the option with minimal RMSE.
For gp-IRT, in both the binary and continuous case, we fit our IRT models using variational inference over 2000 epochs with learning rate 0.1 for the Adam optimizer.

Experiments were all performed on CPU, usually parallelised with 64 workers, with representative execution times (without parallelisation) presented in \cref{fig:stability_and_difficulty_combined}.

\subsection{Binary experiments}\label{app:binary_exp_details}

For our binary-scored experiments in \cref{sec:binary_experiments} we use results processed by \citet{zhang_how_2025} who use two open-source benchmarks.
The first is Open LLM Leaderboard \citep{fourrier_open_2024}, reporting results for 448 models on seven datasets:
\begin{itemize}
\item \textbf{IFEval} \citep{zhou_instruction-following_2023}
\item \textbf{OpenLLM-Math} \citep{hendrycks_measuring_2021-1} (Filtered to only contain the most difficult questions (Level 5).)
\item \textbf{MMLU-Pro} \citep{wang_mmlu-pro_2024}
\item \textbf{ARC-Challenge} \citep{clark_think_2018}
\item \textbf{BBH} \citep{suzgun_challenging_2023}
\item \textbf{GPQA} \citep{rein_gpqa_2024}
\item \textbf{MuSR} \citep{sprague_musr_2024}
\end{itemize}
The second leaderboard comes from HELM (Holistic Evaluation of Language Models) \citep{liang_holistic_2023}, containing results from 83 models on six datasets:
\begin{itemize}
\item \textbf{CommonsenseQA} \citep{talmor_commonsenseqa_2019}
\item \textbf{GSM8K} \citep{cobbe_training_2021}
\item \textbf{LegalBench} \citep{guha_legalbench_2023}
\item \textbf{MATH} \citep{hendrycks_measuring_2021-1}
\item \textbf{Med-QA} \citep{jin_what_2021}
\item \textbf{MMLU} \citep{hendrycks_measuring_2021}
\end{itemize}

\subsection{Continuous experiments}\label{app:cont_exp_details}
For continuous non-pass@$k$ experiments we use eval data provided by \citet{balkir_confident_2026}, on two document summarization datasets: BiolaySumm \citep{xiao_overview_2025}---for which we have ROUGE-L, BERTScore and FKGL metrics---and GovReport \citep{huang_efficient_2021}---for which we have ROUGE-L and BERTScore. 
Although \citet{balkir_confident_2026} provide results from more datasets, we find that the spread of model performance on the other benchmarks is either too small or too sparse to give sensible training sets for our stratified sampling split.
On these datasets we have 22 LLMs each evaluated at four temperatures $\{0.0, 0.4, 0.7, 1.0\}$.


\subsection{\texorpdfstring{Pass@$k$}{Pass@k} experiments}\label{app:pass@k_exp_details}
In \cref{sec:pass@k_experiments} we use pass@$k$ results for $k \in \{1,2,4,8,16,32,64\}$ using 10 LLMs at temperatures $\{0.3,0.6,0.9\}$ using 128 generations\footnote{We don't consider pass@128 since this is extremely high variance with only 128 generations.} per question on four Python benchmarks:
\begin{itemize}
    \item MBPP \citep{austin_program_2021}
    \item MBPP+ \citep{liu_is_2023}
    \item LBPP \citep{matton_leakage_2024}
    \item HumanEval \citep{chen_evaluating_2021}
\end{itemize}
We use the following LLMs to obtain a range of performances:
\begin{itemize}
\item \texttt{GLM 4.7}
\item \texttt{Kimi k2.5}
\item \texttt{Gemma3 1B Base}
\item \texttt{Gemma3 4B Base}
\item \texttt{Olmo3-1125-32B}
\item \texttt{Olmo3-1025-7B}
\item \texttt{Qwen3 4B Base}
\item \texttt{Qwen3 30B A3B Base}
\item \texttt{Qwen3.5 2B Base}
\item \texttt{Qwen3.5 35B A3B Base}
\end{itemize}
Finally, we discard model-temperature combinations in which more than 1\% of samples failed to produce all 128 generations per question, leaving 24 combinations for MBPP+ and HumanEval and 25 for MBPP and LBPP.
For these remaining combinations, we also disregard any benchmark questions for which any combination did not successfully produce all 128 generations.

\section{Mutual information estimators}\label{mrmr:app:mi_estimators}

With discrete data, estimating mutual information can be done using empirical proportions, as we showed for the binary case in \cref{eq:mi-discrete}, \cref{eq:binary_joint} and \cref{eq:binary_marginals}.
However, if one or both of your random variables is continuous, then more sophisticated MI estimators are required.

\subsection{KSG estimator}
First consider the case where we have two continuous random variables, $X$ and $Y$ on sample spaces $\Omega_X$ and $\Omega_Y$ respectively, with joint distribution $p(x, y)$ and marginal distributions $p(x)$ and $p(y)$.
As discussed in \cref{sec:mrmr}, the mutual information between $X$ and $Y$ can be expressed either as a KL-divergence between their joint distribution and the outer-product of their marginals,
\begin{equation}
    \label{mrmr:eq:mi_kl_app}
    I(X,Y) = D_{KL}(P_{X,Y}\; || \; P_X\otimes P_Y )= \int_{\Omega_X} \int_{\Omega_Y} p(x,y)\log\left(\frac{p(x,y)}{p(x)p(y)}\right)dydx \geq 0,
\end{equation}
or as a difference of entropies,
\begin{equation}
    \label{mrmr:eq:mi_entropy_app}
    I(X,Y) = H(X) + H(Y) - H(X,Y) = H(X) - H(X \mid Y) = H(Y) - H(Y \mid X),
\end{equation}
where
\begin{align}
    \label{mrmr:eq:entropies}
    H(X) &= -\int_{\Omega_X} p(x)\log(p(x))dx, \quad H(Y) = -\int_{\Omega_Y} p(y)\log(p(y))dy, \\
    H(X,Y) &= -\int_{\Omega_X} \int_{\Omega_Y} p(x,y)\log(p(x,y))dxdy. \nonumber
\end{align}

A naive approach to calculating $I(X,Y)$ might be to discretize the spaces $\Omega_X$ and $\Omega_Y$ according to some fixed bin size, and then use the empirical joint and marginal distributions to estimate the mutual information.
However, this approach relies heavily on the choice of bin size, and is not robust in the few data regime.
The KSG estimator \citep{kraskov_estimating_2004} instead uses nearest-neighbour distances in the joint space $(X,Y)$ to estimate the joint distribution $p(x,y)$ and marginals, effectively using an adaptive bin size.

Specifically, let us suppose we have $M$ joint samples $z_i = (x_i, y_i) \in \Omega_X \times \Omega_Y$ for $i \in [M]$.
For a given joint sample $z_i = (x_i, y_i) \in \Omega_X \times \Omega_Y$, we first calculate the nearest-neighbour distance $d_i$ to the $k$-th nearest neighbour in the joint space under an $L_\infty$ norm (i.e. the max-norm), and construct a square box centred around $z_i$ with side length $2d_i$.
By definition, the box will contain $k$ points excluding $z_i$ itself. 
Next we define $m_x(i)$ as the number of datapoints whose $x$-coordinate is within distance $d_i$ of $x_i$, and $m_y(i)$ as the number of datapoints whose $y$-coordinate is within distance $d_i$ of $y_i$, in both cases excluding $z_i$ itself:
\begin{equation}
    \label{mrmr:eq:ksg_counts}
    m_x(i) = \sum_{j \in [M]\setminus \{i\}} \mathbb{I}[|x_j - x_i| \leq d_i], \quad m_y(i) = \sum_{j \in [M]\setminus \{i\}} \mathbb{I}[|y_j - y_i| \leq d_i].
\end{equation}

The estimator follows from the entropy decomposition $I(X,Y) = H(X) + H(Y) - H(X,Y)$ (\cref{mrmr:eq:mi_entropy_app}), with each entropy estimated by the nearest-neighbour (Kozachenko--Leonenko) estimator \citep{kozachenko1987sample}: the joint entropy $H(X,Y)$ from the $k$-th neighbour in the joint space (the $\psi(k)$ term) and the marginals $H(X)$ and $H(Y)$ from the counts $m_x(i)$ and $m_y(i)$ within radius $d_i$ (the $\psi(m_x(i)+1)$ and $\psi(m_y(i)+1)$ terms). Reusing the single joint radius $d_i$ for both marginal counts causes the estimators' biases to largely cancel, giving:
\begin{equation}
    \label{mrmr:eq:ksg_estimator}
    \hat{I}_{\text{KSG}}(X,Y) = \psi(k) + \psi(M) - \frac{1}{M}\sum_{i=1}^M (\psi(m_x(i) + 1) + \psi(m_y(i) + 1)),
\end{equation}
where $\psi(x)$ is the digamma function, $\psi(x) = \frac{d}{dx} \log \Gamma(x)$. 
Note that we add 1 to the counts of $m_x(i)$ and $m_y(i)$ to account for the point $z_i$ itself.

This estimator enjoys a number of desirable properties, such as being non-parametric and consistent \citep{gao_demystifying_2018}, and although it does require a choice of $k$, typically this hyperparameter isn't overly sensitive: in practice it is commonly chosen as an integer between roughly 3 and 9.
Using a smaller $k$ is not only faster, but also leads to smaller bias, however, a larger $k$ reduces variance.
We tune $k$ in our experiments over $\{3,4,5,6,7,8,9\}$ and report the results in \cref{app:mrmr_objs}.

\subsection{PCA-corrected KSG}
In \cref{app:mrmr_objs} we find that the PCA-corrected KSG estimator proposed in \citet{gao_efficient_2015} leads to significantly improved performance over the standard KSG estimator.
The authors note that relying on the max-norm box in \cref{mrmr:eq:ksg_counts}, which has volume $V(i) = 4d_i^2$ for datapoint $z_i$, assumes local uniform density within the box, which is unlikely to be true if $X$ and $Y$ are strongly dependent.
Instead, the authors use local PCA on the points in the $k$-neighbourhood of $z_i$ (centred at $z_i$) to find a smaller, rotated rectangle which contains these $k$ points.
The volume of this smaller rectangle is denoted $\bar{V}(i)$ and can be calculated by transforming the $k$ neighbours to new coordinates defined by the principal components and taking the product of the maximal differences in axis values.

Making the weaker assumption of local uniformity within this smaller rectangle, we calculate the PCA-corrected MI estimator as:
\begin{equation}
    \label{mrmr:eq:pca_ksg}
    \hat{I}_{\text{PCA}}(X,Y) = \hat{I}_{\text{KSG}}(X, Y) - \frac{1}{M} \sum_{i=1}^M \log \frac{\bar{V}(i)}{V(i)}.
\end{equation}

\subsection{Discrete-continuous KSG}
For binary-scored benchmark questions, the mRMR relevance terms (\cref{eq:rel_red}) require us to calculate the mutual information between a discrete random variable (the per-question scores, $s(m, q) \in \{0,1\}$) and a continuous random variable (the summary scores, $\bar{s}(m, \mathcal{D}) \in [0,1]$).
To do this, we employ the discrete-continuous MI estimator described by \citet{ross_mutual_2014}.
This estimator works near-identically to the standard KSG estimator, except we modify the nearest-neighbour search to take place in subsets of the data with identical discrete values.

Specifically, we assume now that $X$ is a discrete random variable taking values in some discrete set $\Omega_X$, and $Y$ is a continuous random variable with sample space $\Omega_Y$.
Again, we assume a fixed $k \in \mathbb{N}$ and $M$ joint samples in a set denoted by $\mathcal{Z} = \{z_i = (x_i, y_i) \in \Omega_X \times \Omega_Y : i \in [M]\}$.
For a single datapoint $z_i = (x_i, y_i)$, we denote by $\mathcal{X}_i$ the set of datapoints other than $z_i$ whose discrete values equal $x_i$:
\begin{equation}
    \label{mrmr:eq:discrete_continuous_ksg_set}
    \mathcal{X}_i = \{z_j = (x_j, y_j) \in \mathcal{Z}\setminus \{z_i\} : x_j = x_i\}.
\end{equation}
Then, we define $d_i$ to be the distance between $y_i$ and the $k$-th nearest neighbour in $\mathcal{X}_i$ (in terms of $Y$ values), which allows us to define the set $\mathcal{Z}_i$ of all points (excluding $z_i$) whose $Y$ value is within distance $d_i$ from $y_i$ (inclusive of the $k$-th neighbour itself, which lies at distance exactly $d_i$):
\begin{equation}
    \label{mrmr:eq:discrete_continuous_ksg_subset}
    \mathcal{Z}_i = \{z_j = (x_j, y_j) \in \mathcal{Z}\setminus \{z_i\} : |y_j - y_i| \leq d_i\}.
\end{equation}
We set $m_x(i) = |\mathcal{X}_i|$ and $m_z(i) = |\mathcal{Z}_i|$.
Note that the elements of $\mathcal{Z}_i$ need not have the same discrete value as $x_i$, and so if all $y_j$ values are distinct, then the $k$-th neighbour among $\mathcal{X}_i$ will be the $m_z(i)$-th neighbour among $\mathcal{Z}_i$.

Because $X$ is now discrete, we cannot apply the continuous nearest-neighbour entropy estimator to $H(X)$ or $H(X,Y)$ as we did for the KSG estimator. Instead we use the equivalent decomposition $I(X,Y) = H(Y) - H(Y \mid X)$ and estimate both of these continuous-in-$Y$ entropies with the same estimator as before: $H(Y)$ from neighbours in the full sample (the $\psi(M)$ and $\psi(m_z(i))$ terms) and $H(Y \mid X)$ from neighbours sharing the discrete value $x_i$ (the $\psi(k)$ and $\psi(m_x(i)+1)$ terms), again reusing the shared radius $d_i$ so the biases largely cancel:
\begin{equation}
    \label{mrmr:eq:ross_ksg}
    \hat{I}_\text{Discrete-Continuous}(X, Y) = \psi(M) + \psi(k) - \frac{1}{M}\sum_{i=1}^M (\psi(m_x(i) + 1) + \psi(m_z(i))).
\end{equation}
Note firstly that we add 1 to $m_x(i)$ but not $m_z(i)$: the term $\psi(m_x(i)+1)$ requires the total number of same-discrete-value points \emph{including} $z_i$ itself (as needed for the conditional-entropy estimate), whereas $m_z(i)$ enters as a count of neighbours, which excludes $z_i$ by construction.
Secondly, note that $k$ has a slightly different interpretation here compared to the standard KSG estimator in \cref{mrmr:eq:ksg_estimator}.
In the standard KSG estimator, $k$ is the number of nearest neighbours to consider in the joint space $(X,Y)$ out of our entire dataset of $M$ points, whereas here $k$ is the number of nearest neighbours to consider in each $\mathcal{X}_i$, and in particular, we require that $k \leq |\mathcal{X}_i|$ for each $i$ (otherwise we return 0).
Therefore, in general, we might expect that the optimal value of $k$ will be smaller in the discrete-continuous KSG estimator than in the standard KSG estimator.
This is reflected in our experiments, where we observed optimal performance with $k=5$ in the binary experiments and $k=8$ in the continuous experiments.

\newpage

\section{mRMR ablations}\label{app:mrmr_ablations}

\subsection{mRMR objectives and mutual information estimators}\label{app:mrmr_objs}
Here  we report experiment ablations on both MIQ and MID versions of the mRMR+/++ methods as we vary the nearest-neighbours $k$ hyperparameter in the discrete-continuous MI estimator \citep{ross_mutual_2014}, the KSG MI estimator \citep{kraskov_estimating_2004} and the PCA-corrected KSG estimator \cite{gao_efficient_2015}.
Additionally, we also show results for two further mRMR feature importance functions $g$ proposed in \citep{ding_minimum_2005} specifically for continuous data.
These are the F-test correlation difference (FCD) and the F-test correlation quotient (FCQ) with importance functions:
\begin{align}
    \text{(FC Difference)} \quad g^\text{FCD}(X_i; \C_{(t-1)}, Y) =& F(X_i, Y) - \frac{1}{|\C_{(t-1)}|}\sum_{X_j  \in \mathcal{C}_{(t-1)}}\rho(X_j, X_i), \label{eq:fcd} \\ 
    \text{(FC Quotient)} \quad g^\text{FCQ}(X_i; \C_{(t-1)}, Y) =& \left.F(X_i, Y) \middle/ \left(\frac{1}{|\C_{(t-1)}|}\sum_{X_j \in \mathcal{C}_{(t-1)}}\rho(X_j, X_i)\right),\right.
    \label{eq:fcq}
\end{align}
where $\rho(X_j, X_i)$ is the Pearson correlation and $F(X_i,Y)$ is the F-statistic.

\begin{figure}[!htbp]
    \centering
    \includegraphics[width=\textwidth]{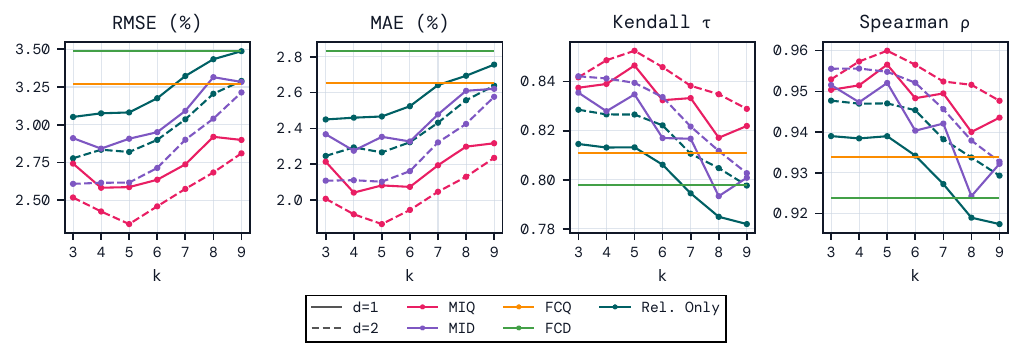}
    \caption{Ablations on mRMR schemes and MI-estimator nearest-neighbour hyperparameter $k$ on binary experiments from \cref{sec:binary_experiments}. FCQ and FCD schemes are also shown as horizontal lines since they don't have this hyperparameter $k$.}
    \label{fig:mrmr_obj_ablation_binary}
\end{figure}

\begin{figure}[!htbp]
    \centering
    \includegraphics[width=\textwidth]{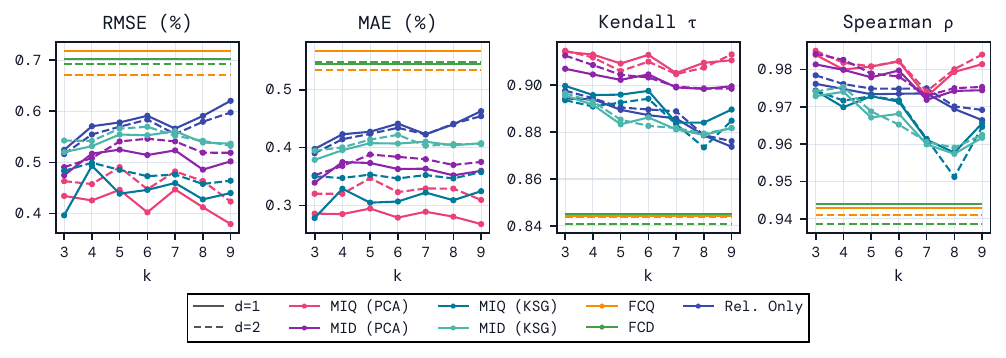}
    \caption{Ablations on mRMR schemes and MI-estimator nearest-neighbour hyperparameter $k$ on continuous non-pass@$k$ experiments from \cref{sec:continuous_experiments}. FCQ and FCD schemes are also shown as horizontal lines since they don't have this hyperparameter $k$.}
    \label{fig:mrmr_obj_ablation_continuous}
\end{figure}

\begin{figure}[!htbp]
    \centering
    \includegraphics[width=\textwidth]{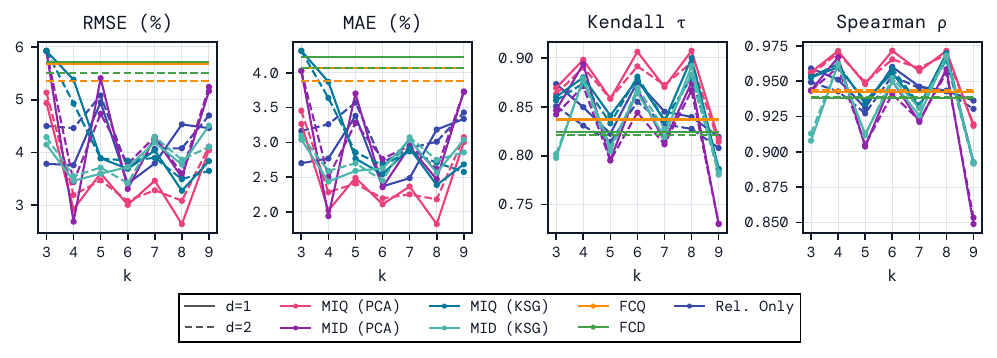}
    \caption{Ablations on mRMR schemes and MI-estimator nearest-neighbour hyperparameter $k$ on pass@$k$ experiments from \cref{sec:pass@k_experiments}. FCQ and FCD schemes are also shown as horizontal lines since they don't have this hyperparameter $k$.}
    \label{fig:mrmr_obj_ablation_pass@k}
\end{figure}

In \cref{fig:mrmr_obj_ablation_binary} we report ablations on binary experiments from \cref{sec:binary_experiments}, finding optimal $k=5$ with $d=2$ on the MIQ variant, with performance degrading significantly in both MIQ and MID for $k > 7$.
In \cref{fig:mrmr_obj_ablation_continuous} we report ablations on continuous non-pass@$k$ experiments from \cref{sec:continuous_experiments} finding overall optimal $k=8$ with $d=2$ on the MIQ scheme using PCA-corrected MI estimators.
Finally, in \cref{fig:mrmr_obj_ablation_pass@k} we report ablations on pass@$k$ experiments from \cref{sec:pass@k_experiments}, finding optimal results with $k=8$ on the MIQ scheme with $d=2$ again using the PCA-corrected MI estimator.
In all settings, we find the FCD and FCQ methods tend to underperform relative to the MI-based methods, except when MI estimates become poor with $k$ values that are too large. 
We also find that coresets that only maximise the MI-calculated relevance (shown as 'Rel. Only') tend to underperform the methods which also take into account the redundancy of a coreset.

\newpage
\subsection{Relevance targets}\label{app:relevance_target}
\begin{figure}[!htbp]
    \centering
    \includegraphics[width=\textwidth]{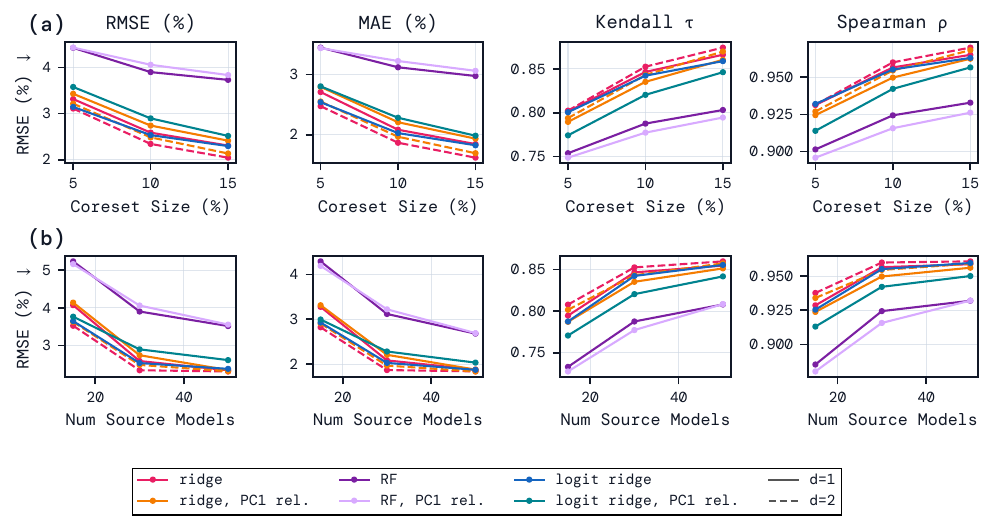}
    \caption{Ablations on binary experiments over the form of prediction (ridge regression, polynomial ridge regression, random-forests, or inverse-logit-transformed ridge/kernel ridge regression) and relevance target $Y_m$ (standard model mean score $\bar{s}(m, \mathcal{D})$ or first-principal component coordinate. 
    }
    \label{fig:mrmr_rel_pred_ablation_binary}
\end{figure}

We also show ablations on the form of prediction applied to mRMR-constructed coresets, using not just the (kernel) ridge regression shown in \cref{sec:experiments}, but also random forests and (kernel) ridge applied to inverse-logit transformed full benchmark scores, $\bar{s}$.
Furthermore, we also consider the effect of replacing $Y_j = \bar{s}(m_j, \mathcal{D})$ in the mRMR relevance calculation (\textit{not} in the regression)  for model $m$ on dataset $\mathcal{D}$ with the dot product between the models' score-vector
\begin{equation}
    \vec{s}_m = [s(m_j, q_1), \ldots, s(m_j, q_N)] \in \mathbb{R}^N,
\end{equation}
and the first principle component of the matrix formed by stacking these vectors across all models,
\begin{equation}
    [\vec{s}_1^T, \ldots, \vec{s}_m^T]^T \in \mathbb{R}^{m \times N}.
\end{equation}
We tried this in order to identify whether a broader relevance signal combining information across models might result in better coreset generation, however, in \cref{fig:mrmr_rel_pred_ablation_binary} we show that for the binary experiments, it's clear than polynomial ridge regression ($d=2$) with the standard relevance target outperforms the other variants.

\newpage
\section{Relevance and redundancy}\label{app:coreset_rel_red}
\begin{figure}[!htbp]
    \centering
    \includegraphics[width=\textwidth]{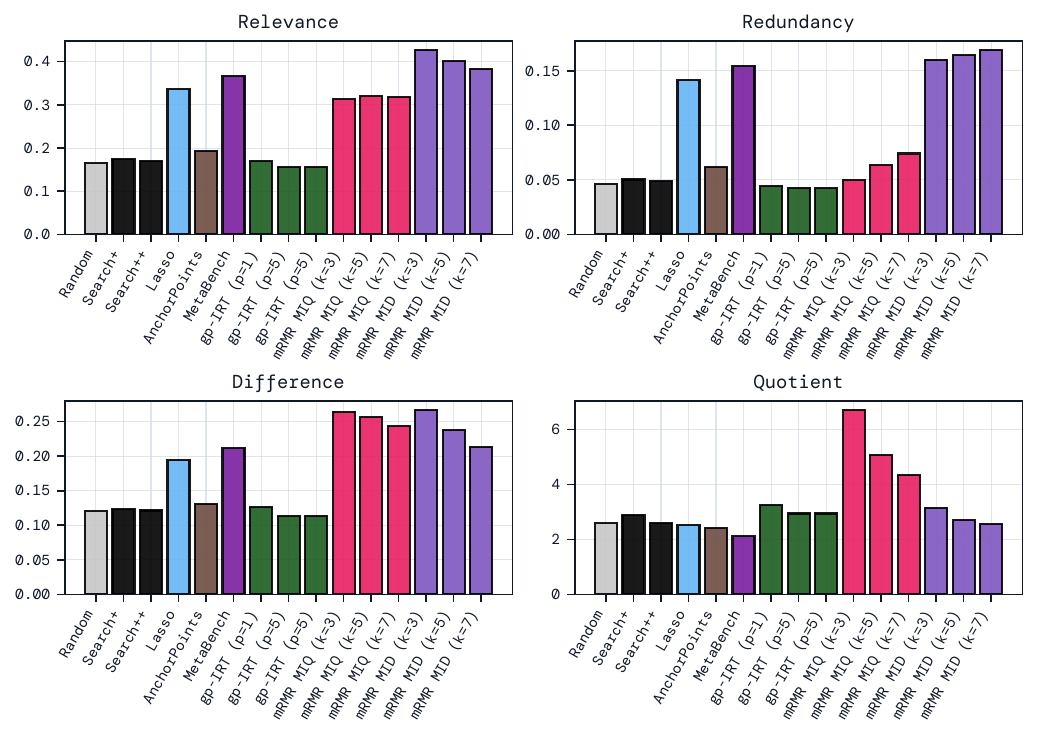}
    \caption{The relevance, redundancy, difference (relevance $-$ redundancy) and quotient (relevance $/$ redundancy) of coresets generated with different methods on binary data.}
    \label{fig:all_methods_rel_red}
\end{figure}

We can obtain much better estimates of mutual information if we use \textit{all} models, not just the source models, in equations \cref{eq:mi-discrete}, \cref{eq:binary_joint} and \cref{eq:binary_marginals}. 
Doing this for the binary case (where MI estimation is most robust) allows us to more accurately estimate the relevance and redundancy (\cref{eq:rel_red}) of generated coresets.
In \cref{fig:all_methods_rel_red} we report the relevance, redundancy, difference (relevance $-$ redundancy) and quotient (relevance $/$ redundancy) of all coreset construction methods.
We find that indeed the greedy algorithm with the MIQ feature importance function $g$ (\cref{eq:miq}) is indeed the best method for maximising the quotient, and in fact is also maximises the difference to about the same extent as the MID function (\cref{eq:mid}).
Whilst MID-based mRMR, lasso and Metabench also achieve relatively high MID scores, their high redundancy leaves them with low-diversity coresets and consequently relatively low quotient scores.

\begin{figure}[!tbp]
    \centering
    \includegraphics[width=\textwidth]{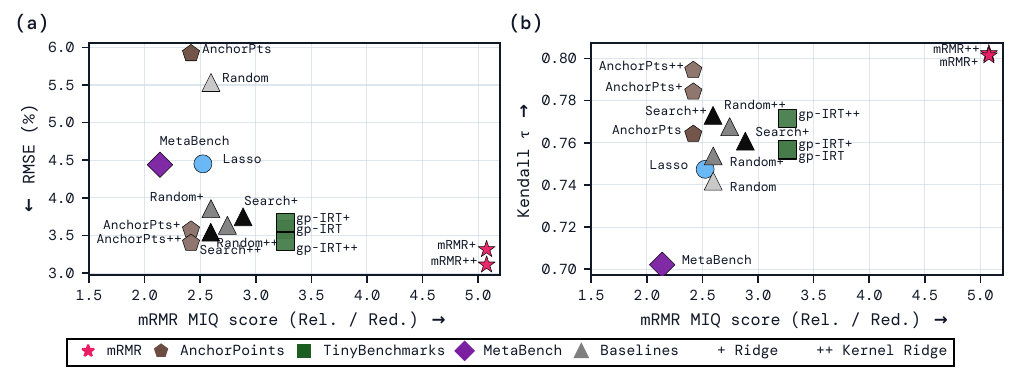}
    \caption{Our method maximises the MIQ mRMR objective (\cref{eq:miq}) better than other methods, leading to reduced prediction error (RMSE; \textbf{a}) and increased ranking correlation (Kendall $\tau$; \textbf{b}). (Same experiment settings as shown in \cref{fig:pareto_main} \textbf{(a)}.)
    }
    \label{fig:pareto_miq}
\end{figure}

In \cref{fig:pareto_miq} we reproduce the results from \cref{fig:pareto_main} \textbf{(a)} with MIQ score on the x-axis, to emphasise that the MIQ-based mRMR methods achieve low RMSE (\cref{fig:pareto_miq} \textbf{(a)}) and high Kendall $\tau$ (\cref{fig:pareto_miq} \textbf{(b)}).

In order to visualise the greedy algorithm from \cref{app:mrmr_algo}, we show the feature importance function $g$ in both its MID and MIQ variants (\cref{eq:mid} and \cref{eq:miq}), as well as the constituent partial relevance and redundancy terms (which are \textit{not} estimates of the full relevance and redundancy in \cref{eq:rel_red}) during coreset construction for 5\% coresets with $M=30$ source models averaged over all binary datasets and random seeds (cut off at the final step for the smallest coreset).

\begin{figure}[!htbp]
    \centering
    \includegraphics[width=\textwidth]{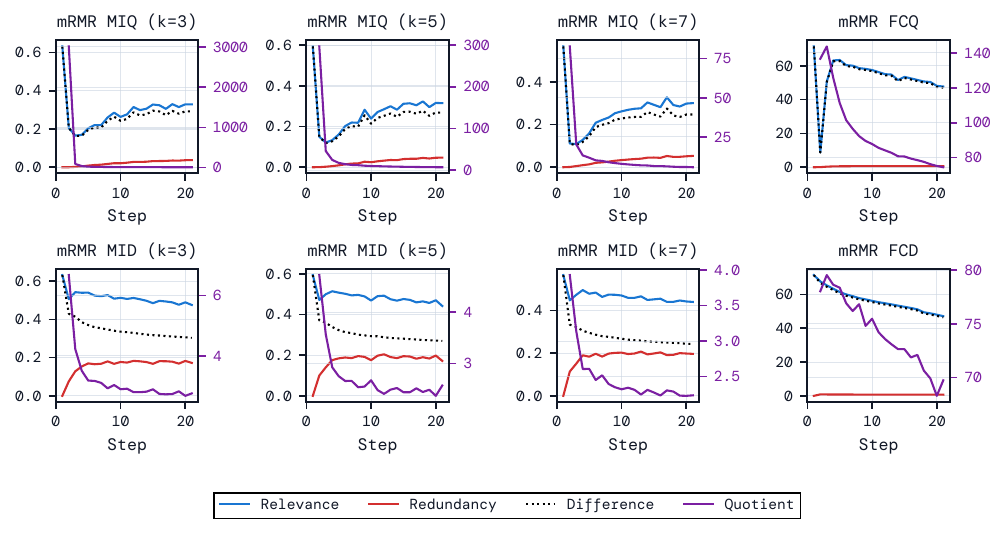}
    \caption{Feature selection metrics during coreset construction for different mRMR variants.}
    \label{fig:mrmr_selection}
\end{figure}

\newpage
\section{IRT details}\label{app:irt_ablations}

\subsection{gp-IRT heuristic}\label{app:tinybench_details}
In \cref{eq:gpirt}, $\lambda = \hat{b}^2/(\hat{b}^2 + \hat{\sigma}^2/4n)$ is calculated using the sample variance of all question scores across all source models, $\hat{\sigma}^2$, and an approximation of the IRT prediction bias, $\hat{b}$, estimated by the mean absolute difference between true scores in a validation set (containing half the source-model data) and predicted scores from an IRT model trained on the other half of source-model data.

\subsection{IRT models specification}\label{app:cirt}
The binary IRT model in \cref{eq:irt} is implemented with the following priors and hyperpriors for a given dimension $d \in \mathbb{N}$ with model $m$ obtaining score $Y_{m,j}$ on question $j$. 
Note that the two continuous models only admit $d=1$.

\begin{align*}
    \text{Hyperpriors:} \quad & \mu_{\beta}, \mu_{\theta, d}, \mu_{\gamma, d} \sim \mathcal{N}(0, 10) \\
    & u_{\beta}, u_{\theta, d}, u_{\gamma, d} \sim \text{Gamma}(1, 1) \\
    \text{Priors:} \quad & \beta_m \sim \mathcal{N}(\mu_{\beta}, u_{\beta}^{-1}) \\
    & \theta_{id} \sim \mathcal{N}(\mu_{\theta, d}, u_{\theta, d}^{-1}) \\
    & \gamma_{md} \sim \mathcal{N}(\mu_{\gamma, d}, u_{\gamma, d}^{-1}) \\
    \text{Likelihood:} \quad & z_{im} = \sum_{d=1}^D (\gamma_{md} \theta_{id}) - \beta_m \\
    & Y_{im} \sim \text{Bernoulli}(\text{logit} = z_{im})
\end{align*}

In sections \cref{sec:continuous_experiments} and \cref{sec:pass@k_experiments} we report gp-IRT results using a Beta IRT model \citep{noel_beta_2007} in place of the standard binary 2PL model from \cref{eq:irt}. 
Here we describe this IRT model, as well as three alternatives that we implemented but found to lead to slightly worse results (although LEGO-IRT \citep{liao_toward_2025} is often very close in performance and sometimes even better).
Each of the models aims to construct a different parameterisation that mirrors the 2PL likelihood in \cref{eq:irt}.

\paragraph{Beta IRT} \citep{noel_beta_2007}
\begin{align*}
    \text{Hyperpriors:} \quad & \mu_{\beta}, \mu_{\theta, d}, \mu_{\gamma, d} \sim \mathcal{N}(0, 10) \\
    & u_{\beta}, u_{\theta, d}, u_{\gamma, d} \sim \text{Gamma}(1, 1) \\
    \text{Priors:} \quad & \beta_m \sim \mathcal{N}(\mu_{\beta}, u_{\beta}^{-1}) \\
    & \theta_{id} \sim \mathcal{N}(\mu_{\theta, d}, u_{\theta, d}^{-1}) \\
    & \gamma_{md} \sim \mathcal{N}(\mu_{\gamma, d}, u_{\gamma, d}^{-1}) \\
    \text{Linear Predictor:} \quad & z_{im} = \sum_{d=1}^D (\gamma_{md} \theta_{id}) - \beta_m \\
    \text{Likelihood:} \quad & \mu_{im} = \exp(z_{im} / 2) \\
    & Y_{im} \sim \text{Beta}(\mu_{im} + 10^{-6}, \mu_{im}^{-1} + 10^{-6})
\end{align*}

\paragraph{LEGO-IRT} \citep{liao_toward_2025}

\begin{align*}
    \text{Hyperpriors:} \quad & \mu_{\beta} \sim \mathcal{N}(0, 10), \quad u_{\beta} \sim \text{Gamma}(1, 1) \\
    & \mu_{\theta, d} \sim \mathcal{N}(0, 10), \quad u_{\theta, d} \sim \text{Gamma}(1, 1) \\
    & \mu_{\gamma, d} \sim \mathcal{N}(0, 10), \quad u_{\gamma, d} \sim \text{Gamma}(1, 1) \\
    \text{Priors:} \quad & \beta_m \sim \mathcal{N}(\mu_{\beta}, u_{\beta}^{-1}) \\
    & \theta_{id} \sim \mathcal{N}(\mu_{\theta, d}, u_{\theta, d}^{-1}) \\
    & \gamma_{md} \sim \mathcal{N}(\mu_{\gamma, d}, u_{\gamma, d}^{-1}) \\
    & \sigma_m \sim \text{HalfNormal}(1) \\
    \text{Linear Predictor:} \quad & \eta_{im} = \sum_{d=1}^D (\gamma_{md} \theta_{id}) - \beta_m \\
    \text{Likelihood:} \quad & \text{logit}(Y_{im}) \sim \mathcal{N}(\eta_{im}, \sigma_m)
\end{align*}

\paragraph{Heteroskedastic Gaussian IRT} \citep{balkir_confident_2026}

\begin{align*}
    \text{Hyperpriors:} \quad & \mu_{\beta}, \mu_{\theta} \sim \mathcal{N}(0, 10), \quad u_{\beta}, u_{\theta} \sim \text{Gamma}(1, 1) \\
    \text{Priors:} \quad & \beta_m \sim \mathcal{N}(\mu_{\beta}, u_{\beta}^{-1}) \\
    & \theta_i \sim \mathcal{N}(\mu_{\theta}, u_{\theta}^{-1}) \\
    & k_m \sim \text{Gamma}(2, 2) \\
    \text{Likelihood:} \quad & \mu_{im} = \sigma(\theta_i - \beta_m) \\
    & \text{var}_{im} = k_m \mu_{im} (1 - \mu_{im}) + 10^{-8} \\
    & Y_{im} \sim \mathcal{N}(\mu_{im}, \sqrt{\text{var}_{im}})
\end{align*}

\paragraph{\texorpdfstring{$\beta^3$ IRT}{Beta-cubed IRT}} \citep{chen_3-irt_2019}
\begin{align*}
    \text{Priors:} \quad & \theta_i \sim \text{Beta}(1, 1) \\
    & \delta_m \sim \text{Beta}(1, 1) \\
    & \alpha_m \sim \mathcal{N}(1, 1) \\
    \text{Parameters:} \quad & \log A_{im} = \alpha_m (\ln \theta_i - \ln \delta_m) \\
    & \log B_{im} = \alpha_m (\ln(1 - \theta_i) - \ln(1 - \delta_m)) \\
    \text{Likelihood:} \quad & A_{im}^* = \exp(\text{clamp}(\log A_{im})) + 10^{-6} \\
    & B_{im}^* = \exp(\text{clamp}(\log B_{im})) + 10^{-6} \\
    & Y_{im} \sim \text{Beta}(A_{im}^*, B_{im}^*)
\end{align*}

\subsection{IRT dimensionality}\label{app:irt_dimension}
\begin{figure}[!htbp]
    \centering
    \includegraphics[width=\textwidth]{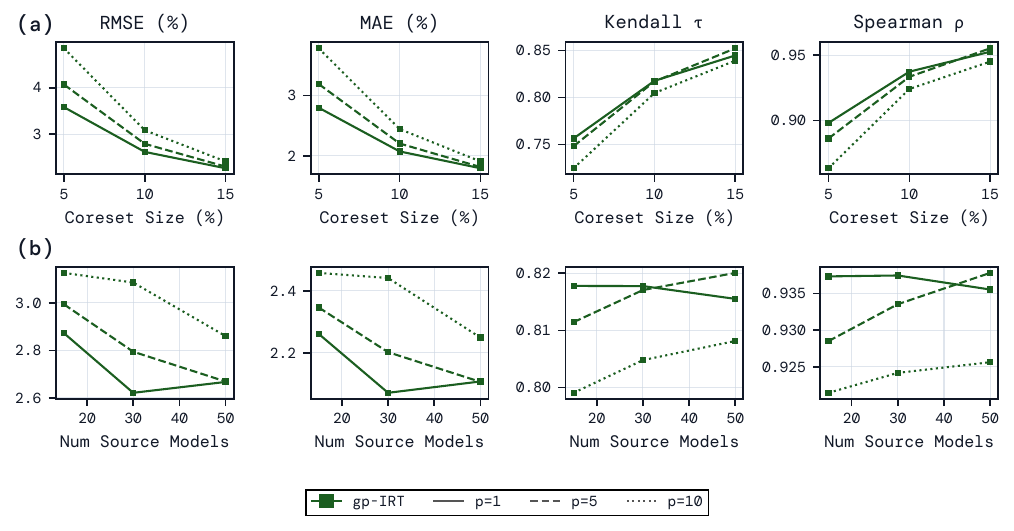}
    \caption{Ablations on IRT dimensionality $p \in \{1,5,10\}$ for gp-IRT on the binary experiments in \cref{sec:binary_experiments}.}
    \label{fig:irt_ablation_binary}
\end{figure}

\begin{figure}[!htbp]
    \centering
    \includegraphics[width=\textwidth]{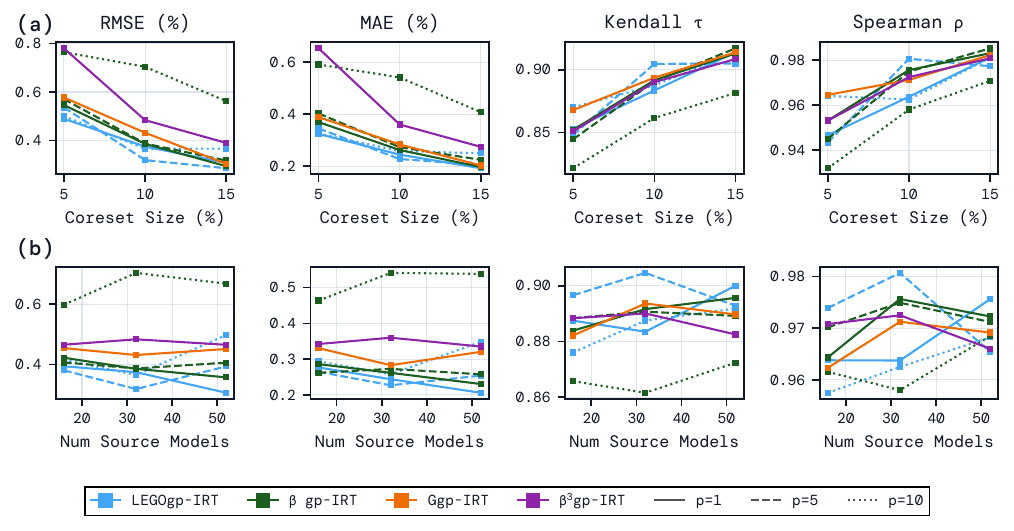}
    \caption{Ablations on IRT dimensionality $p \in \{1,5,10\}$ for gp-IRT variants on the continuous experiments in \cref{sec:pass@k_experiments}.}
    \label{fig:irt_ablation_continuos}
\end{figure}

\begin{figure}[!htbp]
    \centering
    \includegraphics[width=\textwidth]{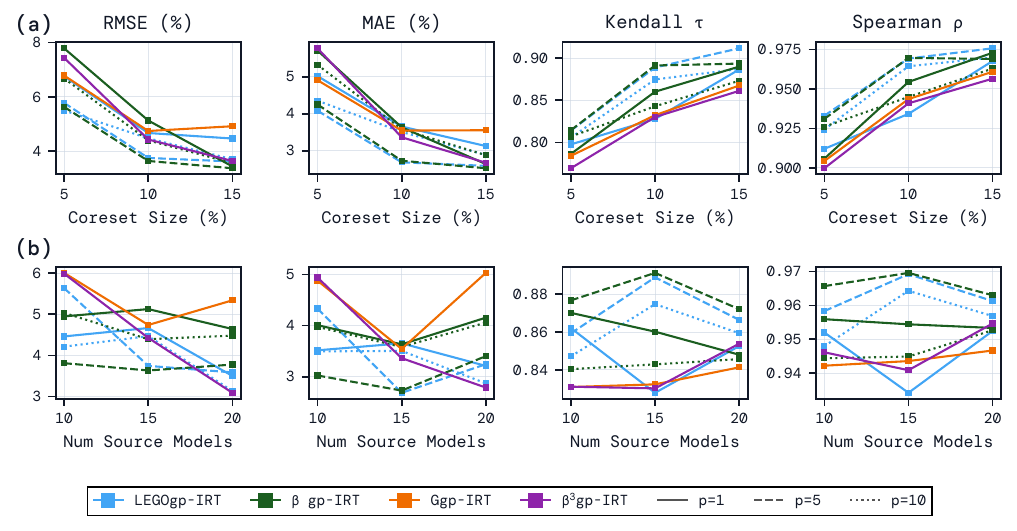}
    \caption{Ablations on IRT dimensionality $p \in \{1,5,10\}$ for gp-IRT variants on the pass@k experiments in \cref{sec:pass@k_experiments}.}
    \label{fig:irt_ablation_pass@k}
\end{figure}

Here we report ablations on the IRT dimensionality $p \in \{1,5,10\}$---values which are suggested in \citet{maia_polo_tinybenchmarks_2024}--for the gp-IRT method.
In particular, \cref{fig:irt_ablation_binary} shows that in the binary setting of \cref{sec:binary_experiments}, optimal results tend to be found with $p=1$, although $p=5$ outperforms this in terms of ranking correlation coefficients for the $M=30$ with 15\% coresets setting (\cref{fig:irt_ablation_binary}(a)) and the $M=50$ with 10\% coreset setting (\cref{fig:irt_ablation_binary}(b)).
In \cref{fig:irt_ablation_continuos} and \cref{fig:irt_ablation_pass@k} we see that most IRT variants lead to fairly similar performance, with $p=5$ dimension $\beta$-IRT and LEGO-IRT achieving similar optimal performance, however, the latter experiences significant instability as as $M$ changes, so we report results on the former in the main text.

\newpage
\section{Varying the number of source models}\label{app:2_row}
\begin{figure}[!htbp]
    \centering
    \includegraphics[width=\textwidth]{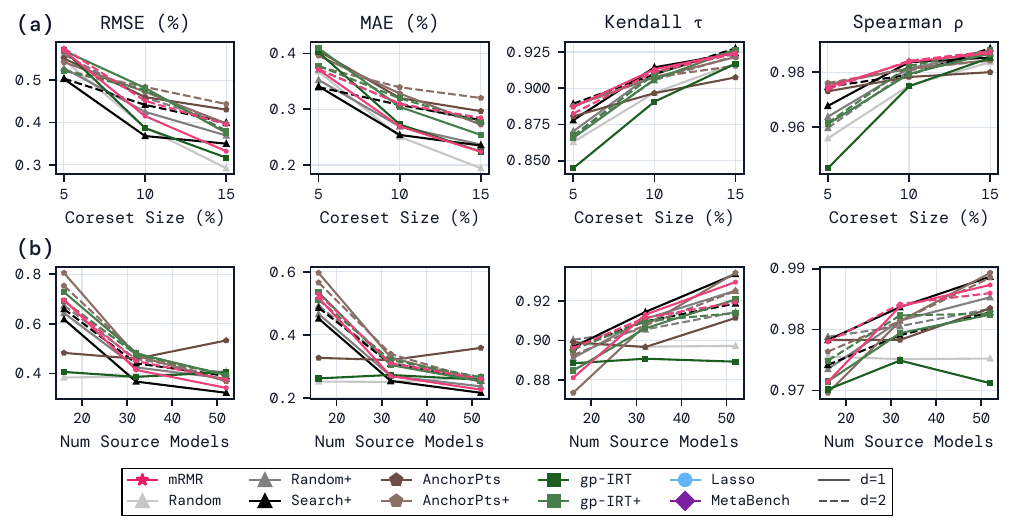}
    \caption{Results from \cref{sec:continuous_experiments}. \textbf{(a)} $M=32$. \textbf{(b)} 10\% coresets.}
    \label{fig:continous_results_2_row}
\end{figure}

\begin{figure}[!htbp]
    \centering
    \includegraphics[width=\textwidth]{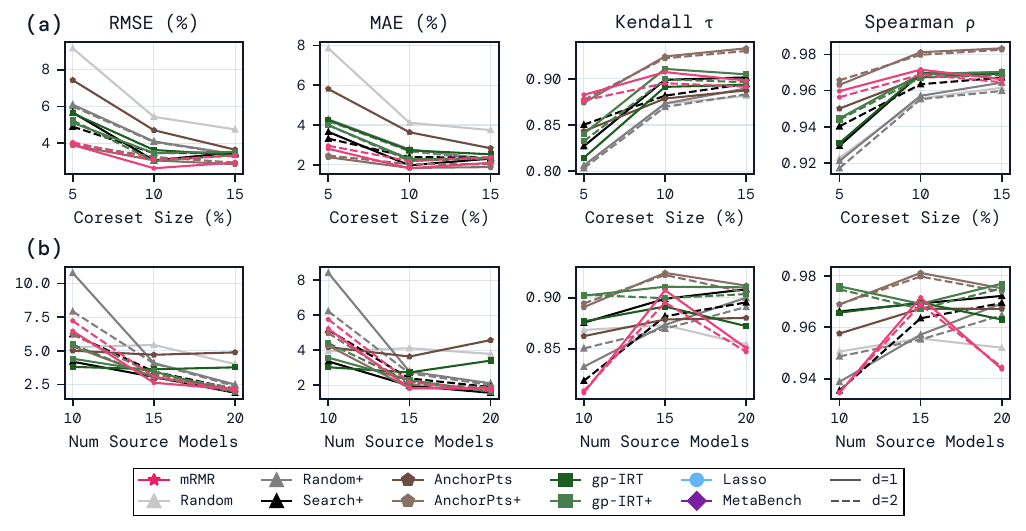}
    \caption{Results from \cref{sec:binary_experiments}. \textbf{(a)} $M=15$. \textbf{(b)} 10\% coresets.}    \label{fig:pass@k_results_2_row}
\end{figure}

In \cref{fig:continuous_results} we report $M=32$ results on continuous non-pass@$k$ experiments from \cref{sec:continuous_experiments} in row \textbf{(a)} and $M=15$ results on pass@$k$ experiments from \cref{sec:pass@k_experiments} in row \textbf{(b)}.
In order to match the information provided for binary experiments in \cref{fig:binary_results}, we show \cref{fig:continous_results_2_row} in which row \textbf{(a)} is the same as \cref{fig:continuous_results} but row \textbf{(b)} shows results on 10\% coresets for varied $M \in \{16,32,52\}$.
Likewise, \cref{fig:pass@k_results_2_row} row \textbf{(a)} reproduces \cref{fig:continuous_results} row \textbf{(b)}, and \cref{fig:pass@k_results_2_row} row \textbf{(b)} shows pass@$k$ experiment results for 10\% coresets with varied $M \in \{10,15,20\}$.

\newpage
\section{Pearson's correlation coefficient results}\label{app:pearson}
\begin{figure}[!htbp]
    \centering
    \includegraphics[width=\textwidth]{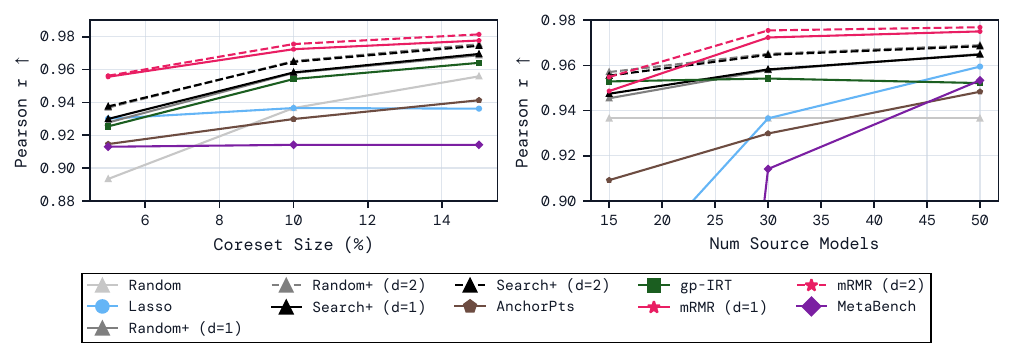}
    \caption{Pearson correlation coefficients on binary experiments in \cref{sec:binary_experiments}.}
    \label{fig:pearson_binary}
\end{figure}

\begin{figure}[!htbp]
    \centering
    \includegraphics[width=\textwidth]{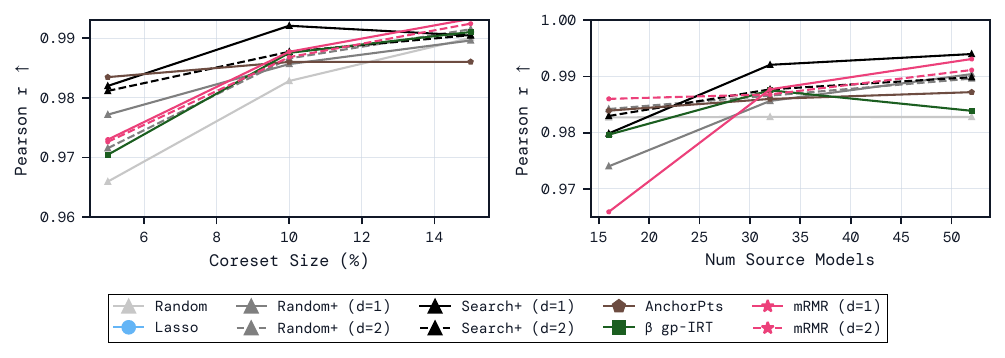}
    \caption{Pearson correlation coefficients on continuous non-pass@$k$ experiments in \cref{sec:continuous_experiments}.}
    \label{fig:pearson_continuous}
\end{figure}

\begin{figure}[!htbp]
    \centering
    \includegraphics[width=\textwidth]{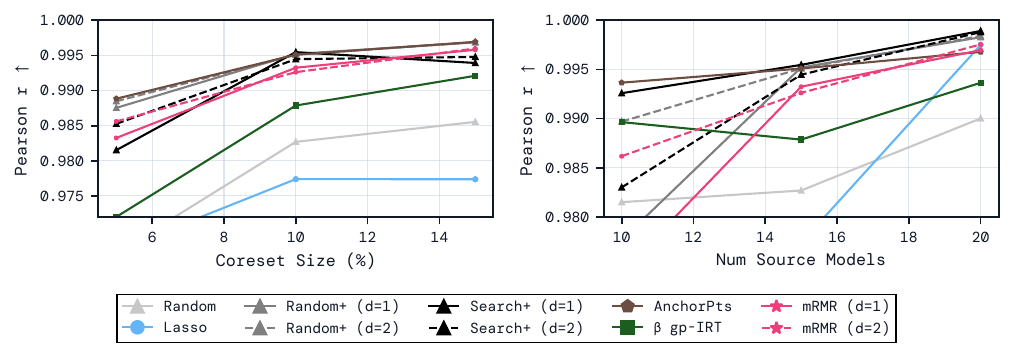}
    \caption{Pearson correlation coefficients on pass@$k$ experiments in \cref{sec:pass@k_experiments}.}
    \label{fig:pearson_pass@k}
\end{figure}

We do not report Pearson correlation coefficients between true and predicted scores of models in the main text because it is more informative to report ranking correlation coefficients $\tau$ and $\rho$ which are not affected by the scale of scores. 
However, in \cref{fig:pearson_binary}, \cref{fig:pearson_continuous} and \cref{fig:pearson_pass@k} we do report these metrics for completion on binary (\cref{sec:binary_experiments}), continuous non-pass@$k$ (\cref{sec:continuous_experiments}), and pass@$k$ (\cref{sec:pass@k_experiments}) respectively.

\newpage
\section{Coreset stability}\label{app:coreset_stability}
\begin{figure}[!htbp]
    \centering
    \includegraphics[width=\textwidth]{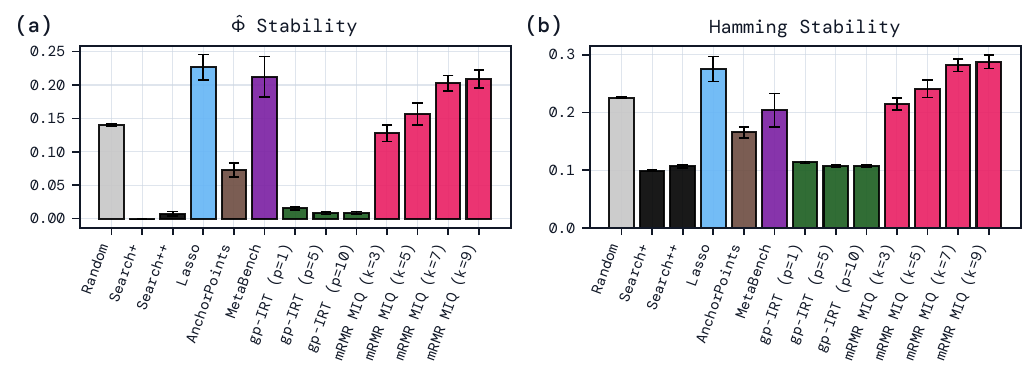}
    \caption{\textbf{(a)} $\hat{\Phi}$ stability on methods in binary experiments, as shown in \cref{fig:stability_and_difficulty_combined}. \textbf{(b)} Hamming stability on the same experiments.}
    \label{fig:coreset_stability}
\end{figure}

In \cref{fig:stability_and_difficulty_combined} \textbf{(b)} we report $\hat{\Phi}$ stability for each method in the binary setting on coresets of size 5\%, 10\% and 15\% with $M \in \{15,30,50\}$. 
This is reproduced in \cref{fig:coreset_stability} \textbf{(a)}, whilst in \cref{fig:coreset_stability} \textbf{(b)} we also report the Hamming stability described in \cref{app:metric_descriptions}, and see much the same behaviour.

\section{Out-of-distribution performance}\label{app:ood}
Although we focus on the in-distribution setting throughout this paper, where our source models well-represent the target models, here we present results on out-of-distribution performance, defined in two ways.
Firstly, in \cref{app:ood_extrap} we investigate the effect of testing on the highest performing 30\% of models (per dataset) whilst training on either the bottom 50\% or bottom 70\%.
Secondly, in \cref{app:ood_new_family} we examine the effect of testing on a new family of models for which no members form part of the source model pool. 

We compare several baseline methods as well as the `augmented inverse propensity weighting' (AIPW) method, proposed by \citet{zhang_how_2025} exactly for this OOD setting.
Based on prediction powered inference (PPI) methods, AIPW uses a coreset, $\mathcal{C}$, of $n$ questions sampled at random from dataset $\mathcal{D}$ of $N$ questions.
For a given test model $\mtest$, AIPW trains a ridge regression model, $g: \mathbb{R}^M \to \mathbb{R}$, to predict the test model's scores $s(\mtest, q)$ using source model scores $[s(m, q)]_{m \in [M]}$ as the features, using the coreset questions as the training data $q \in \mathcal{C}$.
The AIPW estimator uses this model as a proxy to estimate the full-dataset mean, $\bar{s}(\mtest, \mathcal{D})$, and ``debiases'' it as follows:
\begin{align}\label{mrmr:eq:aipw}
    f_\mathcal{C}^{\text{AIPW}}(\mtest) = \bar{s}(\mtest, \mathcal{C}) 
    + \frac{1}{1+ \frac{n}{N-n}} \Bigg[
        \frac{1}{N-n} \sum_{q \in \mathcal{D} \setminus \mathcal{C}} &g\left([s(m, q)]_{m \in [M]}\right) \\
        - \frac{1}{n} \sum_{q \in \mathcal{C}} &g\left([s(m, q)]_{m \in [M]}\right)
    \Bigg]. \notag
\end{align}

In the following tables, we show results for openllm-leaderboard binary datasets, using $M=30$ source models to construct 10\% coresets, reporting MAE, RMSE, Kendall $\tau$ and Spearman $\rho$.
We report 95\% confidence intervals over five random seeds, calculated via bootstrapping over seed-dataset combinations (using dataset-aggregated weighting from \cref{app:metric_descriptions} for MAE and RMSE).
Underlined items indicate the best performing method for a given coreset size and metric, whilst bolding indicates any method whose 95\% confidence interval overlaps that of the best performing method.

\subsection{Extrapolation performance}\label{app:ood_extrap}
In this section, we order the source models by true mean performance on each dataset, and sample our source models from either the bottom 50\% or bottom 70\% of models, before testing only on the top 30\%. 
Results for RMSE, MAE, Kendall $\tau$, and Spearman $\rho$ are given in \cref{tab:ood_extrapolation_rmse,tab:ood_extrapolation_mae,tab:ood_extrapolation_tau,tab:ood_extrapolation_rho} respectively.

Immediately it becomes clear that AIPW is the best performing method in this OOD setting across all four metrics, with performance degrading significantly on other methods (although notably less so for gp-IRT).
Furthermore, we see that in this very difficult setting, the simplest baseline, Random, performs better than all non-AIPW methods across all four metrics, and its 95\% confidence interval overlaps AIPW's on all four (indeed, for the two ranking correlations its point estimate lies inside AIPW's interval).
This displays the vast difference between the in-distribution and out-of-distribution settings: all structured-methods (other than AIPW) work by exploiting the in-distribution behaviour of the source models, but when that behaviour is lost, they perform worse than simply taking the mean on a random sample of questions.

Note, however, that AIPW (and, of course, Random) significantly underperforms in the in-distribution setting with which we are most concerned, hence why we exclude it from our main analyses.

\begin{center}
\footnotesize
\setlength{\tabcolsep}{2pt}
\renewcommand{\arraystretch}{1.05}
\setlength{\LTpre}{0pt}
\setlength{\LTpost}{0pt}
\setlength{\LTleft}{\fill}
\setlength{\LTright}{\fill}
\begin{longtable}{>{\raggedright\arraybackslash}p{0.15\textwidth}>{\raggedleft\arraybackslash}p{0.095\textwidth}>{\raggedleft\arraybackslash}p{0.095\textwidth}>{\raggedleft\arraybackslash}p{0.095\textwidth}>{\raggedleft\arraybackslash}p{0.095\textwidth}>{\raggedleft\arraybackslash}p{0.095\textwidth}>{\raggedleft\arraybackslash}p{0.095\textwidth}>{\raggedleft\arraybackslash}p{0.095\textwidth}>{\raggedleft\arraybackslash}p{0.095\textwidth}}
\caption{OOD extrapolation results: RMSE (10\% coreset, M=30, 7 OpenLLM datasets)}\label{tab:ood_extrapolation_rmse}\\
\toprule
Setting & mRMR++ & mRMR+ & Search+ & Search++ & gp-IRT & AnchorPts & Random & AIPW \\
\midrule
\endfirsthead
\toprule
Setting & mRMR++ & mRMR+ & Search+ & Search++ & gp-IRT & AnchorPts & Random & AIPW \\
\midrule
\endhead
\midrule
\multicolumn{9}{r}{\emph{Continued on next page}} \\
\midrule
\endfoot
\bottomrule
\endlastfoot
In-distribution & \underline{\textbf{1.9302}} {\scriptsize $\pm$ 0.048} & 2.2209 {\scriptsize $\pm$ 0.067} & 2.4800 {\scriptsize $\pm$ 0.11} & 2.3056 {\scriptsize $\pm$ 0.099} & 2.3542 {\scriptsize $\pm$ 0.099} & 4.8824 {\scriptsize $\pm$ 0.31} & 3.5375 {\scriptsize $\pm$ 0.22} & 3.0217 {\scriptsize $\pm$ 0.042} \\
Extrap. (50\% train) & 11.3584 {\scriptsize $\pm$ 0.21} & 12.1745 {\scriptsize $\pm$ 0.20} & 11.3169 {\scriptsize $\pm$ 0.22} & 11.1799 {\scriptsize $\pm$ 0.17} & 4.2770 {\scriptsize $\pm$ 0.42} & 11.6894 {\scriptsize $\pm$ 0.73} & \textbf{3.8177} {\scriptsize $\pm$ 0.27} & \underline{\textbf{3.5738}} {\scriptsize $\pm$ 0.17} \\
Easier extrap. (70\% train) & 8.5759 {\scriptsize $\pm$ 0.25} & 9.1358 {\scriptsize $\pm$ 0.23} & 8.3534 {\scriptsize $\pm$ 0.25} & 8.5146 {\scriptsize $\pm$ 0.16} & 4.1037 {\scriptsize $\pm$ 0.22} & 10.1083 {\scriptsize $\pm$ 0.90} & \textbf{3.8449} {\scriptsize $\pm$ 0.25} & \underline{\textbf{3.5234}} {\scriptsize $\pm$ 0.17} \\
\end{longtable}
\end{center}

\begin{center}
\footnotesize
\setlength{\tabcolsep}{2pt}
\renewcommand{\arraystretch}{1.05}
\setlength{\LTpre}{0pt}
\setlength{\LTpost}{0pt}
\setlength{\LTleft}{\fill}
\setlength{\LTright}{\fill}
\begin{longtable}{>{\raggedright\arraybackslash}p{0.15\textwidth}>{\raggedleft\arraybackslash}p{0.095\textwidth}>{\raggedleft\arraybackslash}p{0.095\textwidth}>{\raggedleft\arraybackslash}p{0.095\textwidth}>{\raggedleft\arraybackslash}p{0.095\textwidth}>{\raggedleft\arraybackslash}p{0.095\textwidth}>{\raggedleft\arraybackslash}p{0.095\textwidth}>{\raggedleft\arraybackslash}p{0.095\textwidth}>{\raggedleft\arraybackslash}p{0.095\textwidth}}
\caption{OOD extrapolation results: MAE (10\% coreset, M=30, 7 OpenLLM datasets)}\label{tab:ood_extrapolation_mae}\\
\toprule
Setting & mRMR++ & mRMR+ & Search+ & Search++ & gp-IRT & AnchorPts & Random & AIPW \\
\midrule
\endfirsthead
\toprule
Setting & mRMR++ & mRMR+ & Search+ & Search++ & gp-IRT & AnchorPts & Random & AIPW \\
\midrule
\endhead
\midrule
\multicolumn{9}{r}{\emph{Continued on next page}} \\
\midrule
\endfoot
\bottomrule
\endlastfoot
In-distribution & \underline{\textbf{1.5373}} {\scriptsize $\pm$ 0.041} & 1.7922 {\scriptsize $\pm$ 0.063} & 1.9331 {\scriptsize $\pm$ 0.092} & 1.8621 {\scriptsize $\pm$ 0.096} & 1.8504 {\scriptsize $\pm$ 0.075} & 3.7749 {\scriptsize $\pm$ 0.22} & 2.8035 {\scriptsize $\pm$ 0.19} & 2.3770 {\scriptsize $\pm$ 0.031} \\
Extrap. (50\% train) & 10.5372 {\scriptsize $\pm$ 0.23} & 11.3458 {\scriptsize $\pm$ 0.21} & 10.4615 {\scriptsize $\pm$ 0.23} & 10.3756 {\scriptsize $\pm$ 0.18} & 3.6700 {\scriptsize $\pm$ 0.42} & 9.0375 {\scriptsize $\pm$ 0.67} & \textbf{3.1216} {\scriptsize $\pm$ 0.25} & \underline{\textbf{2.8896}} {\scriptsize $\pm$ 0.15} \\
Easier extrap. (70\% train) & 7.5783 {\scriptsize $\pm$ 0.26} & 8.1336 {\scriptsize $\pm$ 0.24} & 7.3615 {\scriptsize $\pm$ 0.26} & 7.5870 {\scriptsize $\pm$ 0.16} & 3.4331 {\scriptsize $\pm$ 0.21} & 7.5216 {\scriptsize $\pm$ 0.84} & \textbf{3.1462} {\scriptsize $\pm$ 0.23} & \underline{\textbf{2.8548}} {\scriptsize $\pm$ 0.15} \\
\end{longtable}
\end{center}

\begin{center}
\footnotesize
\setlength{\tabcolsep}{2pt}
\renewcommand{\arraystretch}{1.05}
\setlength{\LTpre}{0pt}
\setlength{\LTpost}{0pt}
\setlength{\LTleft}{\fill}
\setlength{\LTright}{\fill}
\begin{longtable}{>{\raggedright\arraybackslash}p{0.15\textwidth}>{\raggedleft\arraybackslash}p{0.095\textwidth}>{\raggedleft\arraybackslash}p{0.095\textwidth}>{\raggedleft\arraybackslash}p{0.095\textwidth}>{\raggedleft\arraybackslash}p{0.095\textwidth}>{\raggedleft\arraybackslash}p{0.095\textwidth}>{\raggedleft\arraybackslash}p{0.095\textwidth}>{\raggedleft\arraybackslash}p{0.095\textwidth}>{\raggedleft\arraybackslash}p{0.095\textwidth}}
\caption{OOD extrapolation results: Kendall $\tau$ (10\% coreset, M=30, 7 OpenLLM datasets)}\label{tab:ood_extrapolation_tau}\\
\toprule
Setting & mRMR++ & mRMR+ & Search+ & Search++ & gp-IRT & AnchorPts & Random & AIPW \\
\midrule
\endfirsthead
\toprule
Setting & mRMR++ & mRMR+ & Search+ & Search++ & gp-IRT & AnchorPts & Random & AIPW \\
\midrule
\endhead
\midrule
\multicolumn{9}{r}{\emph{Continued on next page}} \\
\midrule
\endfoot
\bottomrule
\endlastfoot
In-distribution & \underline{\textbf{0.8351}} {\scriptsize $\pm$ 0.0036} & 0.8227 {\scriptsize $\pm$ 0.0045} & 0.7763 {\scriptsize $\pm$ 0.0070} & 0.8000 {\scriptsize $\pm$ 0.012} & 0.7959 {\scriptsize $\pm$ 0.0094} & 0.7436 {\scriptsize $\pm$ 0.016} & 0.7698 {\scriptsize $\pm$ 0.012} & 0.7699 {\scriptsize $\pm$ 0.0068} \\
Extrap. (50\% train) & 0.5109 {\scriptsize $\pm$ 0.023} & 0.3933 {\scriptsize $\pm$ 0.026} & 0.4616 {\scriptsize $\pm$ 0.022} & 0.5954 {\scriptsize $\pm$ 0.016} & 0.7305 {\scriptsize $\pm$ 0.010} & 0.6865 {\scriptsize $\pm$ 0.016} & \textbf{0.7581} {\scriptsize $\pm$ 0.013} & \underline{\textbf{0.7666}} {\scriptsize $\pm$ 0.014} \\
Easier extrap. (70\% train) & 0.5857 {\scriptsize $\pm$ 0.022} & 0.5123 {\scriptsize $\pm$ 0.024} & 0.5494 {\scriptsize $\pm$ 0.019} & 0.6124 {\scriptsize $\pm$ 0.013} & 0.7077 {\scriptsize $\pm$ 0.0095} & 0.6981 {\scriptsize $\pm$ 0.019} & \textbf{0.7507} {\scriptsize $\pm$ 0.013} & \underline{\textbf{0.7569}} {\scriptsize $\pm$ 0.015} \\
\end{longtable}
\end{center}

\begin{center}
\footnotesize
\setlength{\tabcolsep}{2pt}
\renewcommand{\arraystretch}{1.05}
\setlength{\LTpre}{0pt}
\setlength{\LTpost}{0pt}
\setlength{\LTleft}{\fill}
\setlength{\LTright}{\fill}
\begin{longtable}{>{\raggedright\arraybackslash}p{0.15\textwidth}>{\raggedleft\arraybackslash}p{0.095\textwidth}>{\raggedleft\arraybackslash}p{0.095\textwidth}>{\raggedleft\arraybackslash}p{0.095\textwidth}>{\raggedleft\arraybackslash}p{0.095\textwidth}>{\raggedleft\arraybackslash}p{0.095\textwidth}>{\raggedleft\arraybackslash}p{0.095\textwidth}>{\raggedleft\arraybackslash}p{0.095\textwidth}>{\raggedleft\arraybackslash}p{0.095\textwidth}}
\caption{OOD extrapolation results: Spearman $\rho$ (10\% coreset, M=30, 7 OpenLLM datasets)}\label{tab:ood_extrapolation_rho}\\
\toprule
Setting & mRMR++ & mRMR+ & Search+ & Search++ & gp-IRT & AnchorPts & Random & AIPW \\
\midrule
\endfirsthead
\toprule
Setting & mRMR++ & mRMR+ & Search+ & Search++ & gp-IRT & AnchorPts & Random & AIPW \\
\midrule
\endhead
\midrule
\multicolumn{9}{r}{\emph{Continued on next page}} \\
\midrule
\endfoot
\bottomrule
\endlastfoot
In-distribution & \underline{\textbf{0.9511}} {\scriptsize $\pm$ 0.0024} & 0.9431 {\scriptsize $\pm$ 0.0032} & 0.9056 {\scriptsize $\pm$ 0.0062} & 0.9224 {\scriptsize $\pm$ 0.0081} & 0.9196 {\scriptsize $\pm$ 0.0099} & 0.8709 {\scriptsize $\pm$ 0.019} & 0.8864 {\scriptsize $\pm$ 0.014} & 0.8940 {\scriptsize $\pm$ 0.0076} \\
Extrap. (50\% train) & 0.6485 {\scriptsize $\pm$ 0.027} & 0.5135 {\scriptsize $\pm$ 0.032} & 0.5945 {\scriptsize $\pm$ 0.027} & 0.7517 {\scriptsize $\pm$ 0.017} & 0.8671 {\scriptsize $\pm$ 0.011} & 0.8286 {\scriptsize $\pm$ 0.017} & \textbf{0.8802} {\scriptsize $\pm$ 0.015} & \underline{\textbf{0.8937}} {\scriptsize $\pm$ 0.015} \\
Easier extrap. (70\% train) & 0.7442 {\scriptsize $\pm$ 0.025} & 0.6603 {\scriptsize $\pm$ 0.027} & 0.6964 {\scriptsize $\pm$ 0.023} & 0.7689 {\scriptsize $\pm$ 0.014} & 0.8430 {\scriptsize $\pm$ 0.010} & 0.8402 {\scriptsize $\pm$ 0.023} & \textbf{0.8724} {\scriptsize $\pm$ 0.015} & \underline{\textbf{0.8835}} {\scriptsize $\pm$ 0.016} \\
\end{longtable}
\end{center}

\subsection{New family performance}\label{app:ood_new_family}
Here we re-run our $M=30$, 10\% coreset, openllm-leaderboard experiments (\cref{sec:binary_experiments}), but for each of our five random seeds we hold out one model family for our test set, ensuring none of the family members appear in the train set (which is constructed from the rest of the models using the same stratified sampling as in the main paper). 
The model families (given by their openllm-leaderboard family name) and their respective family size (which can change based on the benchmark) are: qwen (58 or 59 models); google (20 models); ibm-granite (19 models); meta-llama (20 models); o1-ai (19 models).
We omit experiments on the arc-challenge benchmark in this subsection due to significantly fewer extant results for many of the models in these families.

The RMSE, MAE, Kendall $\tau$ and Spearman $\rho$ of this new-family test setting are compared to their in-distribution equivalents (as in \cref{sec:binary_experiments}) in \cref{tab:single_test_family_rmse,tab:single_test_family_mae,tab:single_test_family_tau_test,tab:single_test_family_rho_test}.
We see similar results as in the in-distribution setting---mRMR++ providing the lowest error and highest ranking correlation coefficient---however, there is indeed a slight worsening in performance in almost all methods. 
Importantly, whilst mRMR++ maintains optimality of performance out of the various methods, we do observe other methods, particularly Search++, reaching performance levels whose 95\% bootstrap confidence intervals overlap with that of mRMR++. 
This is likely due to a limiting of test-set performance diversity, and as such we believe that the more diverse evaluations and analyses in \cref{sec:experiments} are more robust for evaluating each method as they would be used in practice.

\begin{center}
\footnotesize
\setlength{\tabcolsep}{2pt}
\renewcommand{\arraystretch}{1.05}
\setlength{\LTpre}{0pt}
\setlength{\LTpost}{0pt}
\setlength{\LTleft}{\fill}
\setlength{\LTright}{\fill}
\begin{longtable}{>{\raggedright\arraybackslash}p{0.15\textwidth}>{\raggedleft\arraybackslash}p{0.095\textwidth}>{\raggedleft\arraybackslash}p{0.095\textwidth}>{\raggedleft\arraybackslash}p{0.095\textwidth}>{\raggedleft\arraybackslash}p{0.095\textwidth}>{\raggedleft\arraybackslash}p{0.095\textwidth}>{\raggedleft\arraybackslash}p{0.095\textwidth}>{\raggedleft\arraybackslash}p{0.095\textwidth}>{\raggedleft\arraybackslash}p{0.095\textwidth}}
\caption{Held-out model family results: RMSE (10\% coreset, M=30, 6 OpenLLM datasets)}\label{tab:single_test_family_rmse}\\
\toprule
Setting & mRMR++ & mRMR+ & Search+ & Search++ & gp-IRT & AnchorPts & Random & AIPW \\
\midrule
\endfirsthead
\toprule
Setting & mRMR++ & mRMR+ & Search+ & Search++ & gp-IRT & AnchorPts & Random & AIPW \\
\midrule
\endhead
\midrule
\multicolumn{9}{r}{\emph{Continued on next page}} \\
\midrule
\endfoot
\bottomrule
\endlastfoot
In-distribution & \underline{\textbf{1.9045}} {\scriptsize $\pm$ 0.053} & 2.2073 {\scriptsize $\pm$ 0.074} & 2.4457 {\scriptsize $\pm$ 0.11} & 2.2777 {\scriptsize $\pm$ 0.11} & 2.3306 {\scriptsize $\pm$ 0.11} & 4.5445 {\scriptsize $\pm$ 0.32} & 3.5113 {\scriptsize $\pm$ 0.21} & 3.0428 {\scriptsize $\pm$ 0.048} \\
Held-out family & \underline{\textbf{1.9213}} {\scriptsize $\pm$ 0.18} & \textbf{2.1135} {\scriptsize $\pm$ 0.20} & 2.5599 {\scriptsize $\pm$ 0.19} & \textbf{2.2789} {\scriptsize $\pm$ 0.16} & 2.6053 {\scriptsize $\pm$ 0.19} & 5.4329 {\scriptsize $\pm$ 0.99} & 3.3438 {\scriptsize $\pm$ 0.25} & 3.0584 {\scriptsize $\pm$ 0.16} \\
\end{longtable}
\end{center}

\begin{center}
\footnotesize
\setlength{\tabcolsep}{2pt}
\renewcommand{\arraystretch}{1.05}
\setlength{\LTpre}{0pt}
\setlength{\LTpost}{0pt}
\setlength{\LTleft}{\fill}
\setlength{\LTright}{\fill}
\begin{longtable}{>{\raggedright\arraybackslash}p{0.15\textwidth}>{\raggedleft\arraybackslash}p{0.095\textwidth}>{\raggedleft\arraybackslash}p{0.095\textwidth}>{\raggedleft\arraybackslash}p{0.095\textwidth}>{\raggedleft\arraybackslash}p{0.095\textwidth}>{\raggedleft\arraybackslash}p{0.095\textwidth}>{\raggedleft\arraybackslash}p{0.095\textwidth}>{\raggedleft\arraybackslash}p{0.095\textwidth}>{\raggedleft\arraybackslash}p{0.095\textwidth}}
\caption{Held-out model family results: MAE (10\% coreset, M=30, 6 OpenLLM datasets)}\label{tab:single_test_family_mae}\\
\toprule
Setting & mRMR++ & mRMR+ & Search+ & Search++ & gp-IRT & AnchorPts & Random & AIPW \\
\midrule
\endfirsthead
\toprule
Setting & mRMR++ & mRMR+ & Search+ & Search++ & gp-IRT & AnchorPts & Random & AIPW \\
\midrule
\endhead
\midrule
\multicolumn{9}{r}{\emph{Continued on next page}} \\
\midrule
\endfoot
\bottomrule
\endlastfoot
In-distribution & \underline{\textbf{1.5183}} {\scriptsize $\pm$ 0.047} & 1.7863 {\scriptsize $\pm$ 0.068} & 1.9042 {\scriptsize $\pm$ 0.090} & 1.8414 {\scriptsize $\pm$ 0.10} & 1.8265 {\scriptsize $\pm$ 0.081} & 3.4003 {\scriptsize $\pm$ 0.19} & 2.7694 {\scriptsize $\pm$ 0.18} & 2.3916 {\scriptsize $\pm$ 0.034} \\
Held-out family & \underline{\textbf{1.5027}} {\scriptsize $\pm$ 0.12} & \textbf{1.6586} {\scriptsize $\pm$ 0.15} & 2.0288 {\scriptsize $\pm$ 0.17} & 1.8738 {\scriptsize $\pm$ 0.14} & 2.0801 {\scriptsize $\pm$ 0.13} & 4.2251 {\scriptsize $\pm$ 0.75} & 2.6811 {\scriptsize $\pm$ 0.21} & 2.4614 {\scriptsize $\pm$ 0.15} \\
\end{longtable}
\end{center}

\begin{center}
\footnotesize
\setlength{\tabcolsep}{2pt}
\renewcommand{\arraystretch}{1.05}
\setlength{\LTpre}{0pt}
\setlength{\LTpost}{0pt}
\setlength{\LTleft}{\fill}
\setlength{\LTright}{\fill}
\begin{longtable}{>{\raggedright\arraybackslash}p{0.15\textwidth}>{\raggedleft\arraybackslash}p{0.095\textwidth}>{\raggedleft\arraybackslash}p{0.095\textwidth}>{\raggedleft\arraybackslash}p{0.095\textwidth}>{\raggedleft\arraybackslash}p{0.095\textwidth}>{\raggedleft\arraybackslash}p{0.095\textwidth}>{\raggedleft\arraybackslash}p{0.095\textwidth}>{\raggedleft\arraybackslash}p{0.095\textwidth}>{\raggedleft\arraybackslash}p{0.095\textwidth}}
\caption{Held-out model family results: Kendall $\tau$ (10\% coreset, M=30, 6 OpenLLM datasets)}\label{tab:single_test_family_tau_test}\\
\toprule
Setting & mRMR++ & mRMR+ & Search+ & Search++ & gp-IRT & AnchorPts & Random & AIPW \\
\midrule
\endfirsthead
\toprule
Setting & mRMR++ & mRMR+ & Search+ & Search++ & gp-IRT & AnchorPts & Random & AIPW \\
\midrule
\endhead
\midrule
\multicolumn{9}{r}{\emph{Continued on next page}} \\
\midrule
\endfoot
\bottomrule
\endlastfoot
In-distribution & \underline{\textbf{0.8135}} {\scriptsize $\pm$ 0.0050} & 0.7992 {\scriptsize $\pm$ 0.0057} & 0.7447 {\scriptsize $\pm$ 0.0086} & 0.7727 {\scriptsize $\pm$ 0.013} & 0.7741 {\scriptsize $\pm$ 0.012} & 0.7139 {\scriptsize $\pm$ 0.019} & 0.7451 {\scriptsize $\pm$ 0.015} & 0.7411 {\scriptsize $\pm$ 0.0088} \\
Held-out family & \underline{\textbf{0.7979}} {\scriptsize $\pm$ 0.025} & \textbf{0.7836} {\scriptsize $\pm$ 0.028} & 0.7276 {\scriptsize $\pm$ 0.047} & \textbf{0.7634} {\scriptsize $\pm$ 0.030} & 0.7384 {\scriptsize $\pm$ 0.024} & 0.6891 {\scriptsize $\pm$ 0.036} & \textbf{0.7405} {\scriptsize $\pm$ 0.035} & 0.7308 {\scriptsize $\pm$ 0.036} \\
\end{longtable}
\end{center}

\begin{center}
\footnotesize
\setlength{\tabcolsep}{2pt}
\renewcommand{\arraystretch}{1.05}
\setlength{\LTpre}{0pt}
\setlength{\LTpost}{0pt}
\setlength{\LTleft}{\fill}
\setlength{\LTright}{\fill}
\begin{longtable}{>{\raggedright\arraybackslash}p{0.15\textwidth}>{\raggedleft\arraybackslash}p{0.095\textwidth}>{\raggedleft\arraybackslash}p{0.095\textwidth}>{\raggedleft\arraybackslash}p{0.095\textwidth}>{\raggedleft\arraybackslash}p{0.095\textwidth}>{\raggedleft\arraybackslash}p{0.095\textwidth}>{\raggedleft\arraybackslash}p{0.095\textwidth}>{\raggedleft\arraybackslash}p{0.095\textwidth}>{\raggedleft\arraybackslash}p{0.095\textwidth}}
\caption{Held-out model family results: Spearman $\rho$ (10\% coreset, M=30, 6 OpenLLM datasets)}\label{tab:single_test_family_rho_test}\\
\toprule
Setting & mRMR++ & mRMR+ & Search+ & Search++ & gp-IRT & AnchorPts & Random & AIPW \\
\midrule
\endfirsthead
\toprule
Setting & mRMR++ & mRMR+ & Search+ & Search++ & gp-IRT & AnchorPts & Random & AIPW \\
\midrule
\endhead
\midrule
\multicolumn{9}{r}{\emph{Continued on next page}} \\
\midrule
\endfoot
\bottomrule
\endlastfoot
In-distribution & \underline{\textbf{0.9394}} {\scriptsize $\pm$ 0.0033} & 0.9295 {\scriptsize $\pm$ 0.0040} & 0.8815 {\scriptsize $\pm$ 0.0081} & 0.9030 {\scriptsize $\pm$ 0.012} & 0.9029 {\scriptsize $\pm$ 0.013} & 0.8439 {\scriptsize $\pm$ 0.022} & 0.8636 {\scriptsize $\pm$ 0.018} & 0.8695 {\scriptsize $\pm$ 0.011} \\
Held-out family & \underline{\textbf{0.9116}} {\scriptsize $\pm$ 0.022} & \textbf{0.9003} {\scriptsize $\pm$ 0.026} & \textbf{0.8466} {\scriptsize $\pm$ 0.049} & \textbf{0.8777} {\scriptsize $\pm$ 0.028} & 0.8584 {\scriptsize $\pm$ 0.020} & 0.8132 {\scriptsize $\pm$ 0.038} & 0.8471 {\scriptsize $\pm$ 0.041} & 0.8415 {\scriptsize $\pm$ 0.040} \\
\end{longtable}
\end{center}

\newpage 

\section{ONEBench/Plackett-Luce Comparison}\label{app:onebench}

A related direction of work to ours is presented by ONEBench \citep{ghosh_onebench_2025}, which aims to rank models by aggregating scores across an open-ended, ever-expanding sample/question pool.
Specifically, the methodology is robust to the situation in which different models are evaluated on different, arbitrary/uncontrolled missing subsets (this is shown to be robust to up to ~95\% missingness). 
Whilst this is a significantly different problem from ours---we only consider fixed/static coresets with fully-observed results on each coreset question, and care about predicted scores, not just rankings---it is worth checking how the suggested approach compares to those investigated in the present work.

As described in the ONEBench paper, we implemented Plackett-Luce as a pairwise comparison model (i.e. the Bradley-Terry model): we build a win/loss matrix for each coreset question, where each coreset question ``votes" for whichever target models it scores correctly over those it scores incorrectly. 
The Plackett-Luce model is then used to recover a ranking from this win/loss matrix. 
Note that we do so using the mRMR-constructed coresets and Search++'s coresets, meaning we're only evaluating Plackett-Luce on the Stage-2 (prediction) step, as it is not designed to select questions.
(ONEBench \textit{does} provide a method for selecting coreset questions using a $k$-nearest neighbours search over question-embeddings (either generated via an LLM or via question metadata), however, this places the method outside of the scope of all other coreset-construction methods we examine in this paper, which require only historic per-question per-source-model scores.)

The results are shown in \cref{tab:onebench}, pooled across all 5 seeds and all 7 openllm-leaderboard benchmarks for $M=30$ and 10\% coresets.
We see that Plackett-Luce does not beat our kernel regression approach, and doesn't beat naive mean-score averaging (which often does surprisingly well at predicted rankings, if not predicted scores).
We believe this is consistent with the scope argument: with no missing data to reconstruct, there's little for Plackett-Luce modelling to add over the raw scores.


\begin{center}
\begin{figure}[h]
\footnotesize
\setlength{\tabcolsep}{2pt}
\renewcommand{\arraystretch}{1.05}
\setlength{\LTpre}{0pt}
\setlength{\LTpost}{0pt}
\setlength{\LTleft}{\fill}
\setlength{\LTright}{\fill}
\begin{longtable}{>{\raggedright\arraybackslash}p{0.17\textwidth}>{\raggedright\arraybackslash}p{0.07\textwidth}>{\raggedleft\arraybackslash}p{0.155\textwidth}>{\raggedleft\arraybackslash}p{0.155\textwidth}>{\raggedleft\arraybackslash}p{0.155\textwidth}>{\raggedleft\arraybackslash}p{0.155\textwidth}}
\caption{Same coreset, four Stage-2 approaches: Kendall $\tau$ / Spearman $\rho$ (M=30, 10\% coreset)}\label{tab:onebench}\\
\toprule
Coreset used & Metric & Ridge regression (d=1) & Kernel ridge regression (d=2) & Naive mean coreset score & Plackett-Luce \\
\midrule
\endfirsthead
\toprule
Coreset used & Metric & Ridge regression (d=1) & Kernel ridge regression (d=2) & Naive mean coreset score & Plackett-Luce \\
\midrule
\endhead
\midrule
\multicolumn{6}{r}{\emph{Continued on next page}} \\
\midrule
\endfoot
\bottomrule
\endlastfoot
mRMR++'s & $\tau$ & 0.8093 {\scriptsize $\pm$ 0.0049} & \textbf{0.8219} {\scriptsize $\pm$ 0.0042} & \underline{\textbf{0.8247}} {\scriptsize $\pm$ 0.0055} & 0.8059 {\scriptsize $\pm$ 0.0061} \\
 & $\rho$ & 0.9357 {\scriptsize $\pm$ 0.0037} & \underline{\textbf{0.9443}} {\scriptsize $\pm$ 0.0028} & \textbf{0.9425} {\scriptsize $\pm$ 0.0034} & 0.9375 {\scriptsize $\pm$ 0.0039} \\
\midrule
Search+'s & $\tau$ & 0.7568 {\scriptsize $\pm$ 0.0071} & \underline{\textbf{0.7785}} {\scriptsize $\pm$ 0.0073} & \textbf{0.7604} {\scriptsize $\pm$ 0.012} & 0.7345 {\scriptsize $\pm$ 0.012} \\
 & $\rho$ & 0.8918 {\scriptsize $\pm$ 0.0068} & \underline{\textbf{0.9091}} {\scriptsize $\pm$ 0.0069} & 0.8788 {\scriptsize $\pm$ 0.014} & 0.8692 {\scriptsize $\pm$ 0.014} \\
 \bottomrule
\end{longtable}

\end{figure}
\end{center}

\newpage

\section{Coreset difficulty}\label{app:coreset_difficulty}
\begin{figure}[!htbp]
    \centering
    \includegraphics[width=\textwidth]{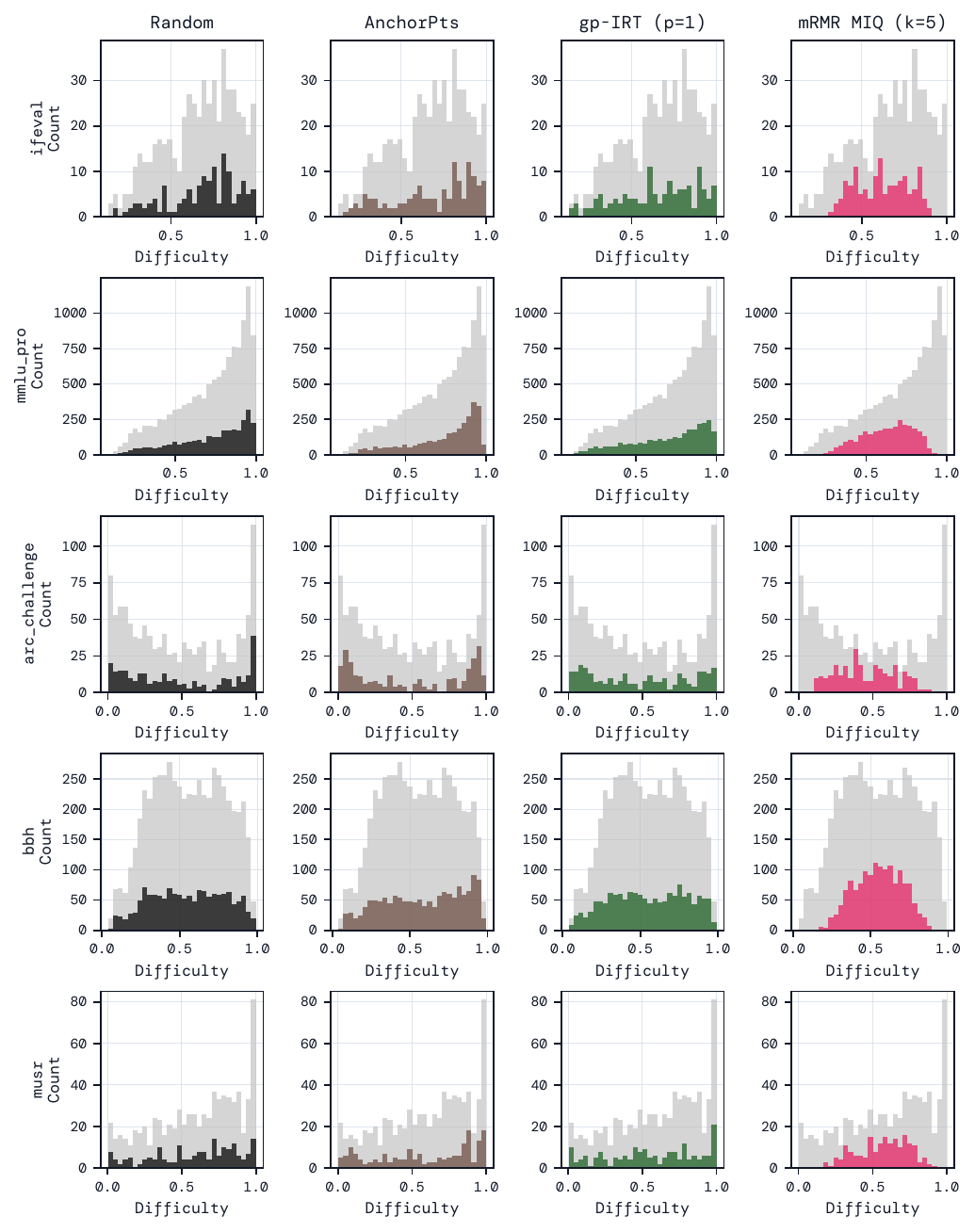}
    \caption{Question difficulty distribution histograms on various datasets for coresets generated using different methods.}
    \label{fig:coreset_difficulty_full}
\end{figure}

In \cref{fig:stability_and_difficulty_combined} \textbf{(a)} we show the distribution of question difficulty in MMLU-Pro \citep{wang_mmlu-pro_2024} and the distribution of question difficulty on coresets generated by random sampling, AnchorPoints, gp-IRT and mRMR.
In \cref{fig:coreset_difficulty_full} we show the same behaviour (mRMR focussing on medium-difficulty questions rather than the more broadly representative questions of the other corsets) on more datasets.

\newpage
\section{Per-dataset true vs predicted comparison}\label{app:true_v_pred}
\begin{figure}[!htbp]
    \centering
    \includegraphics[width=\textwidth]{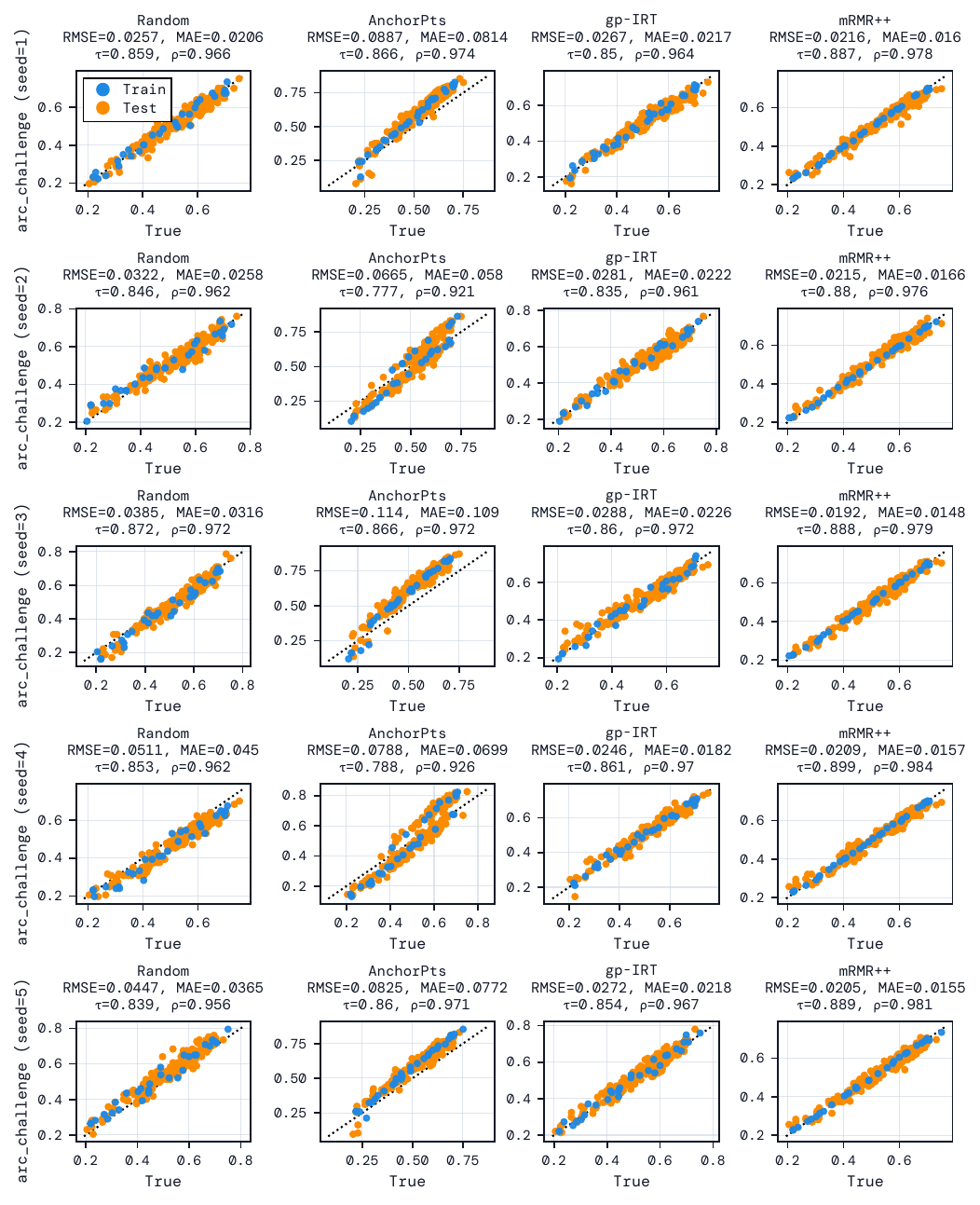}
    \caption{True versus predicted scores on ARC-Challenge, with each row representing a different train-test split via random seed.}
    \label{fig:true_vs_pred_arc-c_per_seed}
\end{figure}

\begin{figure}[!htbp]
    \centering
    \includegraphics[width=\textwidth]{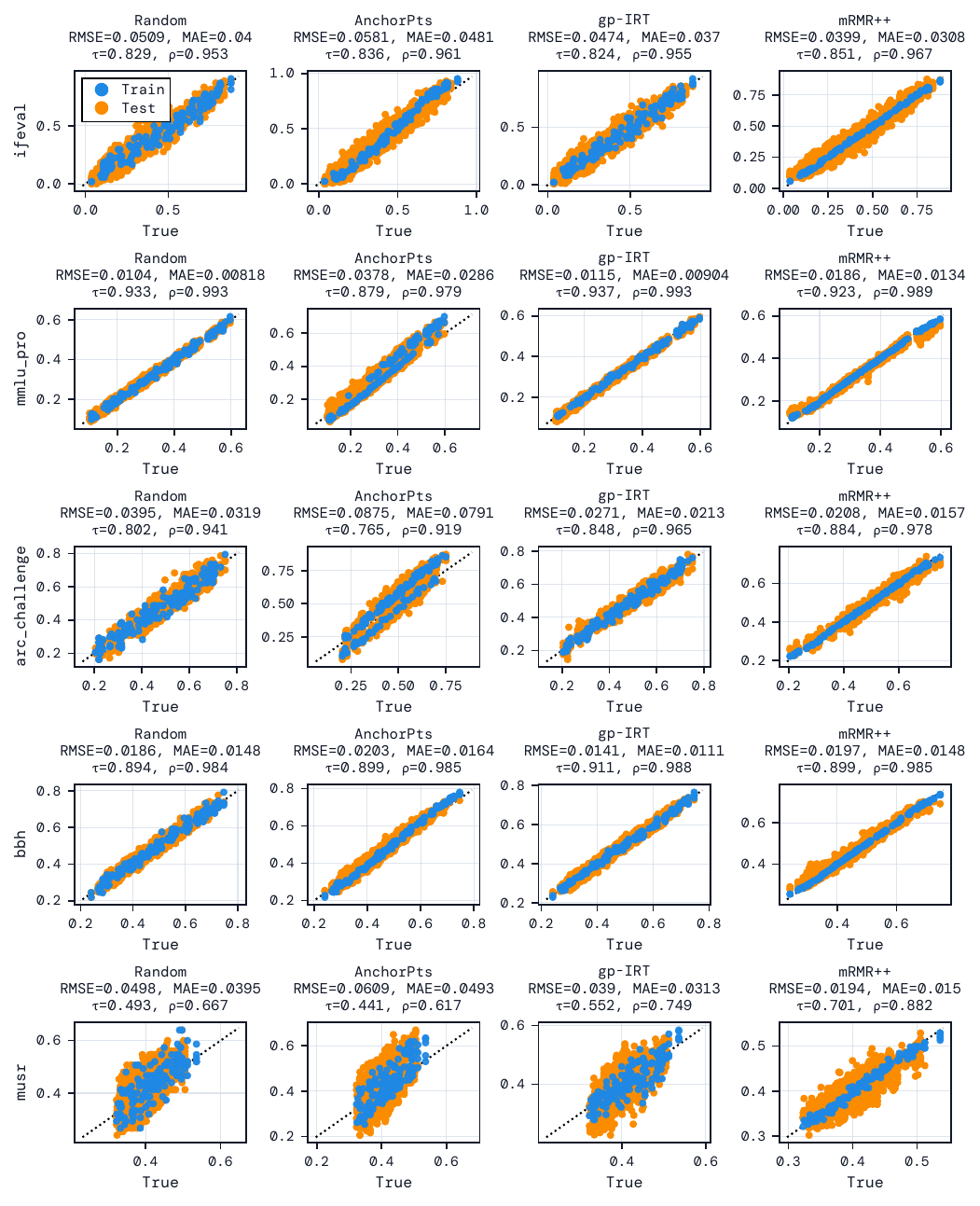}
    \caption{True versus predicted scores on more binary datasets (the same as in \cref{fig:coreset_difficulty_full}) across five random seeds for different train-test splits.}
    \label{fig:true_vs_pred_full}
\end{figure}

In \cref{fig:true_vs_pred_arc-c} we show the true versus predicted scores on the ARC-Challenge \citep{clark_think_2018} dataset from random sampling, AnchorPoints, gp-IRT and mRMR++ across five different random seeds for different training data splits.
To improve the clarity of predictive behaviour, in \cref{fig:true_vs_pred_arc-c_per_seed} we show the same results but split up each seed (from 1 to 5) into different rows.
Additionally, in \cref{fig:true_vs_pred_full} we show the results from \cref{fig:true_vs_pred_arc-c} but with different datasets on each row (the same datasets as in \cref{fig:coreset_difficulty_full}).

\end{document}